%% file: main.tex
\documentclass{article}

\usepackage{microtype}
\usepackage{graphicx}
\usepackage{subcaption}
\usepackage{booktabs} 
\usepackage{makecell}
\usepackage{multirow}
\usepackage{tablefootnote}
\usepackage{pdflscape}
\usepackage{enumitem}
\usepackage{pifont}
\newcommand{\cmark}{\ding{51}}%
\newcommand{\xmark}{\ding{55}}%
\usepackage{lipsum}

\usepackage{hyperref}

\usepackage[accepted]{icml2024}

\usepackage{amsmath}
\usepackage{amssymb}
\usepackage{mathtools}
\usepackage{amsthm}
\usepackage[bb=dsserif]{mathalpha}
\input{math_commands}

\usepackage[capitalize,noabbrev]{cleveref}

\theoremstyle{plain}

\theoremstyle{definition}

\theoremstyle{remark}

\usepackage[textsize=tiny]{todonotes}

\icmltitlerunning{Unified Training of Universal Time Series Forecasting Transformers}

\def\modelname{\textsc{Moirai}}
\def\smallmodel{{\modelname}\textsubscript{Small}}
\def\basemodel{{\modelname}\textsubscript{Base}}
\def\largemodel{{\modelname}\textsubscript{Large}}

\def\lotsa{\texttt{LOTSA}}
\def\libname{\textsc{Uni}\(^2\)\textsc{TS}}

\begin{document}

\twocolumn[
\icmltitle{Unified Training of Universal Time Series Forecasting Transformers}



\icmlsetsymbol{equal}{*}

\begin{icmlauthorlist}
\icmlauthor{Gerald Woo}{sf,smu}
\icmlauthor{Chenghao Liu}{sf}
\icmlauthor{Akshat Kumar}{smu}
\icmlauthor{Caiming Xiong}{sf}
\icmlauthor{Silvio Savarese}{sf}
\icmlauthor{Doyen Sahoo}{sf}
\end{icmlauthorlist}

\icmlaffiliation{sf}{Salesforce AI Research}
\icmlaffiliation{smu}{School of Computing and Information Systems, Singapore Management University}

\icmlcorrespondingauthor{Gerald Woo}{gwoo@salesforce.com}
\icmlcorrespondingauthor{Chenghao Liu}{chenghao.liu@salesforce.com}

\icmlkeywords{Machine Learning, Time Series Forecasting, Large Time Series Model, Universal Forecasting, Large Pre-trained Model, Foundation Model}

\vskip 0.3in
]



\printAffiliationsAndNotice{}  

\input{section/0_abstract}
\input{section/1_introduction}
\input{section/2_related_work}
\input{section/3_0_method}
\input{section/4_0_experiments}
\input{section/5_conclusion}

\section*{Impact Statement}
This paper presents work whose goal is to advance the field of Machine Learning. There are many potential societal consequences of our work, none which we feel must be specifically highlighted here.

\bibliography{main}
\bibliographystyle{icml2024}

\newpage
\input{appendix/appendix}

\end{document}

%% file: math_commands.tex
\usepackage{amsmath,amsfonts,bm,upgreek}

\def\eqref#1{equation~\ref{#1}}

\def\1{\bm{1}}

\def\rh{{\textnormal{h}}}

\def\rl{{\textnormal{l}}}

\def\rt{{\textnormal{t}}}

\def\rx{{\textnormal{x}}}

\def\rvphi{{\bm{\upphi}}}

\def\rmD{{\mathbf{D}}}

\def\rmY{{\mathbf{Y}}}
\def\rmZ{{\mathbf{Z}}}

\def\vtheta{{\bm{\theta}}}
\def\vphi{{\bm{\phi}}}

\def\vx{{\bm{x}}}
\def\vy{{\bm{y}}}
\def\vz{{\bm{z}}}

\def\mD{{\bm{D}}}

\def\mR{{\bm{R}}}

\def\mW{{\bm{W}}}

\def\mY{{\bm{Y}}}
\def\mZ{{\bm{Z}}}

\DeclareMathAlphabet{\mathsfit}{\encodingdefault}{\sfdefault}{m}{sl}
\SetMathAlphabet{\mathsfit}{bold}{\encodingdefault}{\sfdefault}{bx}{n}

\def\gD{{\mathcal{D}}}

\def\gT{{\mathcal{T}}}

\def\emA{{A}}

\def\emE{{E}}

\newcommand{\E}{\mathbb{E}}

\newcommand{\R}{\mathbb{R}}



%% file: section/0_abstract.tex
\begin{abstract}
Deep learning for time series forecasting has traditionally operated within a one-model-per-dataset framework, limiting its potential to leverage the game-changing impact of large pre-trained models. The concept of \textit{universal forecasting}, emerging from pre-training on a vast collection of time series datasets, envisions a single Large Time Series Model capable of addressing diverse downstream forecasting tasks. However, constructing such a model poses unique challenges specific to time series data: \(i)\) cross-frequency learning, \(ii)\) accommodating an arbitrary number of variates for multivariate time series, and \(iii)\) addressing the varying distributional properties inherent in large-scale data.
To address these challenges, we present novel enhancements to the conventional time series Transformer architecture, resulting in our proposed \textbf{M}asked Enc\textbf{\textsc{o}}der-based Un\textbf{\textsc{i}}ve\textbf{\textsc{r}}s\textbf{\textsc{a}}l T\textbf{\textsc{i}}me Series Forecasting Transformer (\textbf{{\modelname}}). Trained on our newly introduced Large-scale Open Time Series Archive ({\lotsa}) featuring over \(27\mathrm{B}\) observations across nine domains, {\modelname} achieves competitive or superior performance as a zero-shot forecaster when compared to full-shot models. Code, data, and model weights can be found at \url{https://github.com/SalesforceAIResearch/uni2ts}.
\looseness=-1
\end{abstract}

%% file: section/1_introduction.tex
\section{Introduction}
In the era of foundation models (FMs) \citep{bommasani2021foundation}, 
the landscape of deep learning for time series forecasting is experiencing a revolution.
In contrast to FMs capable of tackling a multitude of downstream tasks, the current deep forecasting paradigm, involving training a model on a single dataset with a fixed context and prediction length, appears increasingly antiquated, lacking the capacity to generalize or adapt to diverse scenarios or datasets.
Given the unreasonable effectiveness of large pre-trained models in improving performance and data efficiency via transfer learning in modalities like vision and language \citep{dosovitskiy2020vit, brown2020gpt3}, we are starting to see a push to transition away from the existing paradigm, towards a \textit{universal forecasting} paradigm (see \cref{fig:problem}), where a single large pre-trained model is able to handle any time series forecasting problem.
However, the road to building a universal time series forecasting model is mired with challenges. 
\looseness=-1

\input{figure/problem}

Unlike the modalities of vision and language which have the unified formats of images and text respectively, time series data is highly heterogeneous. 
Firstly, the frequency (e.g. minutely, hourly, daily sampling rates) of time series plays an important role in determining the patterns present in the time series. 
Cross-frequency learning has been shown to be a challenging task due to negative interference \citep{van2023cross}, with existing work simply avoiding this problem for multi-frequency datasets by learning one model per frequency \citep{oreshkin2020nbeats}. 
Secondly, time series data are heterogeneous in terms of dimensionality, whereby multivariate time series can have different number of variates. Furthermore, each variate measures a  semantically different quantity across datasets. While considering each variate of a multivariate time series independently \citep{nie2023patchtst,vijay2023tsmixer} can sidestep this problem, we expect a universal model to be sufficiently flexible to consider multivariate interactions and take into account exogenous covariates. 
Thirdly, probabilistic forecasting is a critical feature often required by practitioners, yet, different datasets have differing support and distributional properties -- for example, using a symmetric distribution (e.g. Normal, Student-T) as the predictive distribution is not suitable for positive time series -- making standard approaches of pre-defining a simple parametric distribution \citep{salinas2020deepar} to be insufficiently flexible to capture a wide variety of datasets.
Lastly, a large pre-trained model capable of universal forecasting requires a large-scale dataset from diverse domains. Existing time series datasets are insufficiently large to support the training of such models.
\looseness=-1

Starting from a masked encoder architecture which has been shown to be a strong candidate architecture for scaling up pre-trained time series forecasting models \citep{woo2023pushing}, we alleviate the above issues by introducing novel modifications which allows the architecture to handle the heterogeneity of arbitrary time series data.
Firstly, we propose to learn multiple input and output projection layers to handle the differing patterns from time series of varying frequencies. Using patch-based projections with larger patch sizes for high-frequency data and vice versa, projection layers are specialized to learn the patterns of that frequency.
Secondly, we address the problem of varying dimensionality with our proposed Any-variate Attention, which simultaneously considers both time and variate axes as a single sequence, leveraging Rotary Position Embeddings (RoPE) \citep{su2024roformer}, and learned binary attention biases \citep{yang2022tableformer} to encode time and variate axes respectively. Importantly, Any-variate Attention allows the model to take as input arbitrary number of variates.
Thirdly, we overcome the issue of requiring flexible predictive distributions with a mixture of parametric distributions. Furthermore, optimizing the negative log-likelihood of a flexible distribution has the added benefit of being competitive with target metric optimization \citep{awasthi2022mle}, a powerful feature for pre-training universal forecasters, given that it can be evaluated with any target metric subsequently.
\looseness=-1

\input{table/model_comparison}

To power the training of our Large Time Series Model (LTM), we introduce the Large-scale Open Time Series Archive ({\lotsa}), the largest collection of open time series datasets with \(27\mathrm{B}\) observations across nine domains. We optimize the negative log-likelihood of the mixture distribution, and randomly sample context and prediction lengths during training, allowing for flexible downstream usage of the pre-trained model. 
We train our proposed method, 
\textbf{M}asked Enc\textbf{\textsc{o}}der-based Un\textbf{\textsc{i}}ve\textbf{\textsc{r}}s\textbf{\textsc{a}}l T\textbf{\textsc{i}}me Series Forecasting Transformer 
({\modelname}\footnote{In ancient Greek religion and mythology, the Moirai, often known in English as the Fates, were the personifications of destiny. \citep{wiki2023moirai}}),
in three sizes -- {\smallmodel}, {\basemodel}, and {\largemodel}, with \(14\mathrm{m}\), \(91\mathrm{m}\), and \(311\mathrm{m}\) parameters respectively. We perform experimental evaluations on both in and out-of-distribution settings, and show that {\modelname} consistently achieves competitive or superior performance compared to state-of-the-art full-shot baselines.
Our contributions are summarized as follows: \looseness=-1
\begin{enumerate}
    \item We introduce a novel Transformer architecture to support the requirements of universal forecasting. Crucially, the components we propose extend beyond masked encoders and are versatile, applicable to a broad range of Transformer variants. \looseness=-1
    \item We introduce {\lotsa}, a new large-scale collection of open time series datasets to empower pre-training of LTMs. {\lotsa}, the model weights, and our library for unified training of universal time series models, {\libname}, will be fully open sourced.
    \item Trained on {\lotsa} data, {\modelname} achieves competitive or superior performance as a zero-shot forecaster when compared to full-shot models.
\end{enumerate}

%% file: figure/problem.tex
\begin{figure}[t]
\begin{center}
\centerline{\includegraphics[width=\columnwidth]{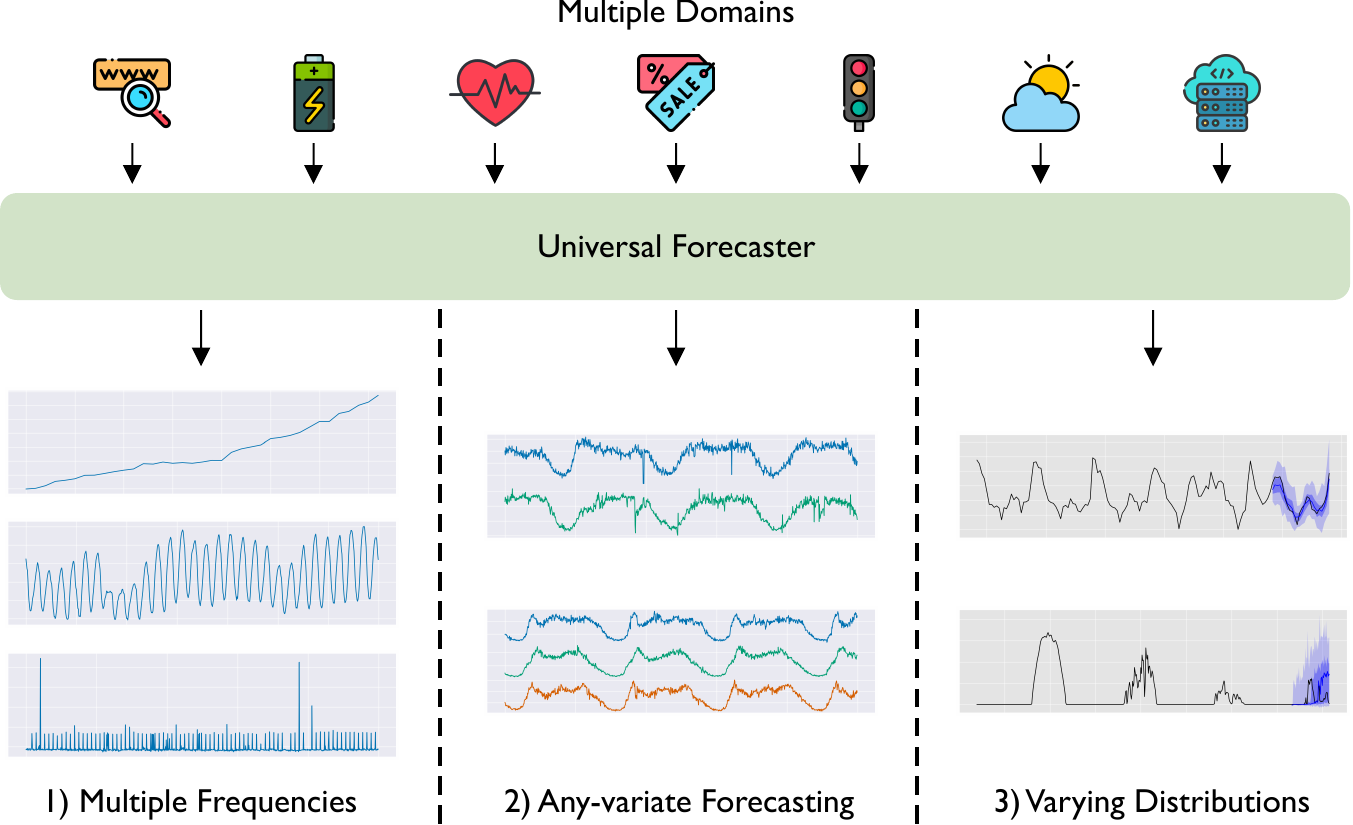}}
\caption{
A \textit{universal forecaster} is a large pre-trained model capable of handling any time series forecasting problem. It is trained on a large-scale time series dataset spanning multiple domains. Compared to the existing paradigm, universal forecasting faces the three key issues of \(i)\) multiple frequencies, \(ii)\) any-variate forecasting, and \(iii)\) varying distributions. 
}
\label{fig:problem}
\end{center}
\vskip -0.3in
\end{figure}

%% file: table/model_comparison.tex
\begin{table}[t]
    \caption{
    Comparison between pre-trained forecasting models. Further discussion on the notion of a ``flexible distribution'' can be found in \cref{app:flexible_distribution}.
    }
    \label{tab:model_comparison}%
    \vskip 0.05in
    \centering
    \resizebox{\columnwidth}{!}{
    \begin{tabular}{lccccc}
        \toprule
             &
            \makecell{\textbf{Any-variate}\\\textbf{(Zero-shot)}} &
            \makecell{\textbf{Probabilistic}\\\textbf{Forecasting}} &
            \makecell{\textbf{Flexible}\\\textbf{Distribution}} &
            \makecell{\textbf{Pre-training Data (Size)}} &
            \textbf{Open-source} \\
        \midrule
            {\modelname} &
            \cmark &
            \cmark &
            \cmark &
            {\lotsa} (\(>27\mathrm{B}\)) &
            \cmark \\

            TimeGPT-1 &
            \cmark &
            \cmark &
            \xmark &
            Unknown (\(100\mathrm{B}\)) &
            \xmark \\

            ForecastPFN &
            \xmark &
            \xmark &
            - &
            Synthetic Data (\(60\mathrm{M}\))&
            \cmark \\

            Lag-Llama &
            \xmark & 
            \cmark &
            \xmark &
            Monash (\(<1\mathrm{B}\)) &
            \cmark \\

            TimesFM &
            \xmark &
            \xmark &
            - &
            Wiki + Trends + Others (\(>100\mathrm{B}\)) &
            \cmark \\

            TTM &
            \xmark &
            \xmark &
            - &
            Monash (\(<1\mathrm{B}\)) &
            \cmark \\

            LLMTime & 
            \xmark &
            \cmark &
            \cmark &
            Web-scale Text &
            \cmark \\
            
        \bottomrule
    \end{tabular}%
}
    \vskip -0.1in
\end{table}%

%% file: section/2_related_work.tex
\section{Related Work}
\label{sec:related_work}
\paragraph{Pre-training for Zero-shot Forecasting}
\cref{tab:model_comparison} provides a summary of the key differences between recent pre-trained models with zero-shot forecasting capabilities, which is a recently emerging field.
TimeGPT-1 \citep{garza2023timegpt} first presented a closed-source model, offering zero-shot forecasting capabilities as well as supporting fine-tuning through an API, currently only available to their beta users. 
ForecastPFN \citep{dooley2023forecastpfn} proposes to pre-train on synthetic time series, which can be subsequently be leveraged as a zero-shot forecaster, albeit specialized for data or time limited settings.
Lag-llama \citep{rasul2023lagllama} works towards a foundation model for time series forecasting, leveraging the LLaMA \citep{touvron2023llama} architecture design with lagged time series features, and also presents neural scaling laws for time series forecasting.
TimesFM \citep{das2023predct} is a patch-based decoder-only foundation model for time series forecasting, introducing a larger output patch size for faster decoding. They collected a massive amount of data from Google Trends and Wiki pageviews to pre-train their model in combination with open-data.
Tiny Time Mixers (TTMs) \citep{ekambaram2024ttms} is a concurrent work leveraging lightweight mixer-style architecture. They perform data augmentation by downsampling high-frequency time series, and support multivariate downstream tasks by fine-tuning an exogenous mixer.
leverage Large Language Models (LLMs), pre-trained on web-scale text data, have been leveraged for zero-shot forecasting. Specifically, LLMTime \citep{gruver2023llmtime} treats time series as strings, applying careful pre-processing based on the specific LLMs' tokenizer, showing that LLMs have the inherent capability to perform zero-shot forecasting.
\looseness=-1

\paragraph{Pre-train + Fine-tune for Time Series Forecasting}
Pre-training with subsequent fine-tuning on downstream forecasting tasks has predated the recent zero-shot forecasting efforts. Denoising autoencoders \citep{zerveas2021transformer} and contrastive learning \citep{yue2022ts2vec,woo2022cost} have been shown to be effective pretext tasks for time series forecasting, but have largely been applied to the existing paradigm of pre-training and fine-tuning on the same dataset, without exploring their generalization capabilities. More recently, \citet{dong2023simmtm} explored combining both reconstruction and contrastive based pre-training approaches, and performed initial explorations into cross-dataset transfer.
The topic has been well explored, and we refer readers to more comprehensive surveys \citep{zhang2023ssl,ma2023survey}.
``Reprogramming'' is a recent direction which involves fine-tuning the model weights of an LLM which has been pre-trained on text data, for downstream tasks for other modalities. \citet{zhou2023gpt4ts,jin2023timellm} introduce modules and fine-tuning methods to adapt LLMs for time series tasks including forecasting. \citet{liu2024unitime} has explored leveraging pre-trained LLMs on the cross-dataset setting.
\looseness=-1

%% file: section/3_0_method.tex
\section{Method}
\paragraph{Problem Formulation}
Consider a dataset of \(N\) time series \(\gD = \{(\mY^{(i)}, \mZ^{(i)})\}_{i=1}^N\), where \(\mY^{(i)} = (\vy^{(i)}_1, \vy^{(i)}_2, \ldots, \vy^{(i)}_{T_i}) \in \R^{d_{y_i} \times T_i}\) is a target time series of \(d_{y_i}\) variates and \(T_i\) time steps. Each time series is associated with a set of covariates \(\mZ^{(i)} = (\vz^{(i)}_1, \vz^{(i)}_2, \ldots, \vz^{(i)}_{T_i}) \in \R^{d_{z_i} \times T_i}\). 
The goal is to forecast the predictive distribution \(p(\mY_{t:t+h}|\vphi)\) by predicting distribution parameters \(\vphi\) via a learned model \(f_{\vtheta}: (\mY_{t-l:t}, \mZ_{t-l:t+h}) \mapsto \hat{\vphi}\) which maximizes the log-likelihood: \looseness=-1
\begin{align}
    & \max_{\vtheta} \underset{\substack{(\rmY, \rmZ) \sim p(\gD)\\ \quad\; (\rt, \rl, \rh) \sim p(\gT|\gD)}}{\E} \log p(\rmY_{\rt:\rt+\rh} | \hat{\rvphi}), \nonumber \\
    & \text{s.t.} \; \hat{\rvphi} = f_{\vtheta}(\rmY_{\rt-\rl:\rt}, \rmZ_{\rt-\rl:\rt+\rh}), \label{eq:optim_problem}
\end{align}
where \(p(\gD)\) is the data distribution which samples for a time series, \((\mY, \mZ)\), and \(p(\gT|\gD)\) is the task distribution which defines the lookback window, \(\mY_{t-l:t} = (\vy_{t-l}, \ldots, \vy_{t-1})\) with context length \(l\) and forecast horizon, \(\mY_{t:t+h} = (\vy_t, \ldots, \vy_{t+h-1})\) with prediction length \(h\). \looseness=-1

\input{section/3_1_method}
\input{section/3_2_method}

%% file: section/3_1_method.tex
\subsection{Architecture}
\input{figure/architecture}
Illustrated in \cref{fig:architecture}, {\modelname} follows a (non-overlapping) patch-based approach to modeling time series with a masked encoder architecture. 
One of our proposed modifications to extend the architecture to the any-variate setting is to ``flatten'' multivariate time series, considering all variates as a single sequence. 
Patches are subsequently projected into vector representations via a multi patch size input projection layer.
The \texttt{[mask]} signifies a learnable embedding which replaces patches falling within the forecast horizon. The output tokens are then decoded via the multi patch size output projection into the parameters of the mixture distribution.
While not visualized, (non-learnable) instance normalization \citep{kim2022reversible} is applied to inputs/outputs, aligning with the current standard practice for deep forecasting models.

The core Transformer module is an encoder-only Transformer architecture, leveraging various improvements as proposed by recent state-of-the-art LLM architectures. We use pre-normalization \citep{xiong2020layer} and replace all LayerNorms with RMSNorm \citep{zhang2019root}, and also apply query-key normalization \citep{henry2020qknorm}. The non-linearity in FFN layers are replaced with SwiGLU \citep{shazeer2020glu}, adjusting the hidden dimension to have equal number of parameters as the original FFN layer. We omit biases in all layers of the Transformer module.

\subsubsection{Multi Patch Size Projection Layers}
In the context of universal forecasting, a single model should possess the capability to handle time series spanning a wide range of frequencies. Existing patch-based architectures rely on a single patch size hyperparameter, a legacy feature from the prevailing one-model-per-dataset paradigm. Instead, we aim for a more flexible strategy: opting for a larger patch size to handle high-frequency data, thereby lower the burden of the quadratic computation cost of attention while maintaining a long context length. Simultaneously, we advocate for a smaller patch size for low-frequency data to transfer computation to the Transformer layers, rather than relying solely on simple linear embedding layers. To implement this approach, we propose learning multiple input and output embedding layers, each associated with varying patch sizes. The selection of the appropriate patch size for a given time series frequency relies on pre-defined settings (see \cref{app:architecture_details_patch_size}). Note that we only learn one set of projection weights per patch size, which is shared amongst frequencies if there is an overlap based on the settings. \looseness=-1

\subsubsection{Any-variate Attention}
Universal forecasters must be equipped to handle arbitrary multivariate time series. Existing time series Transformers often rely on an independent variate assumption or are limited to a single dimensionality due to embedding layers mapping \(\R^{d_y} \to \R^{d_h}\), where \(\R^{d_h}\) is the hidden dimension.
We overcome this limitation as shown in \cref{fig:architecture}, by flattening a multivariate time series to consider all variates as a single sequence. This introduces a new requirement of having variate encodings to enable the model to disambiguate between different variates in the sequence. Furthermore, we need to ensure that permutation equivariance w.r.t. variate ordering, and permutation invariance w.r.t. variate indices are respected. Conventional approaches like sinusoidal or learned embeddings do not meet these requirements, and are unable to handle an arbitrary number of variates. To address this, we propose Any-variate Attention, leveraging binary attention biases to encode variate indices.

Dropping layer and attention head indices, and scaling factor for brevity, the attention score between the \((i,m)\)-th query where \(i\) represents the time index and \(m\) represents the variate index, and the \((j,n)\)-th key, \(\emA_{ij,mn} \in \R\), is given by: \looseness=-1
\begin{align}
    \emE_{ij,mn} & = (\mW^Q \vx_{i,m})^T \mR_{i-j} (\mW^K \vx_{j,n}) \nonumber \\
    & \quad \; + u^{(1)} * \mathbb{1}_{\{m=n\}} + u^{(2)} * \mathbb{1}_{\{m\ne n\}}, \label{eq:attention_pre_softmax} \\
    \emA_{ij,mn} & = \frac{\exp\{\emE_{ij,mn}\}}{\sum_{k,o} \exp\{\emE_{ik,mo}\}}, \label{eq:attention_softmax}
\end{align}
where \(\mW^Q \vx_{i,m}, \mW^K \vx_{j,n} \in \R^{d_h}\) are the respective query and key vectors, \(\mR_{i-j} \in \R^{d_h \times d_h}\) is the rotary matrix \citep{su2024roformer}, \(u^{(1)}, u^{(2)} \in \R\) are learnable scalars for each head in each layer, and
\(
\mathbb{1}_{\{\text{cond}\}} = \left\{\begin{smallmatrix*}[l]
1 \text{, if cond} \\
0 \text{, otherwise}
\end{smallmatrix*}\right.
\) is the indicator function.
The binary attention bias component allows for disambiguation between variates via attention scores, fulfills the criteria of permutation equivariance/invariance w.r.t. variate ordering/indices, and can extend to arbitrary number of variates. \looseness=-1

\subsubsection{Mixture Distribution}
To achieve the goal of having a flexible distribution, yet ensuring that operations of sampling and evaluating the loss function  remains simple, we propose to use a mixture of parametric distributions.
A mixture distribution of \(c\) components has p.d.f.: 
\begin{align}
    p(\rmY_{t:t+h}|\hat{\vphi}) = \sum_{i=1}^c w_i p_i(\rmY_{t:t+h}|\hat{\vphi}_i),
\end{align}
where \(\hat{\vphi} = \{w_1, \hat{\vphi}_1, \ldots, w_c, \hat{\vphi}_c\}\), and \(p_i\) is the \(i\)-th component's p.d.f.
While the choice of mixture components is flexible and implementing any combination of parametric distributions is straightforward, we specifically propose to use the following mixture components: 
\(i)\) a Student's t-distribution which has shown to be a robust option for general time series,
\(ii)\) a negative binomial distribution for positive count data,
\(iii)\) a log-normal distribution to model right-skewed data commonly across economic and and natural phenomena,
and \(iv)\) a low variance normal distribution for high confidence predictions.
Further details can be found in \cref{app:architecture_details_mixture}.
\looseness=-1

%% file: figure/architecture.tex
\begin{figure*}[ht!]
\begin{center}
\centerline{\includegraphics[width=\textwidth]{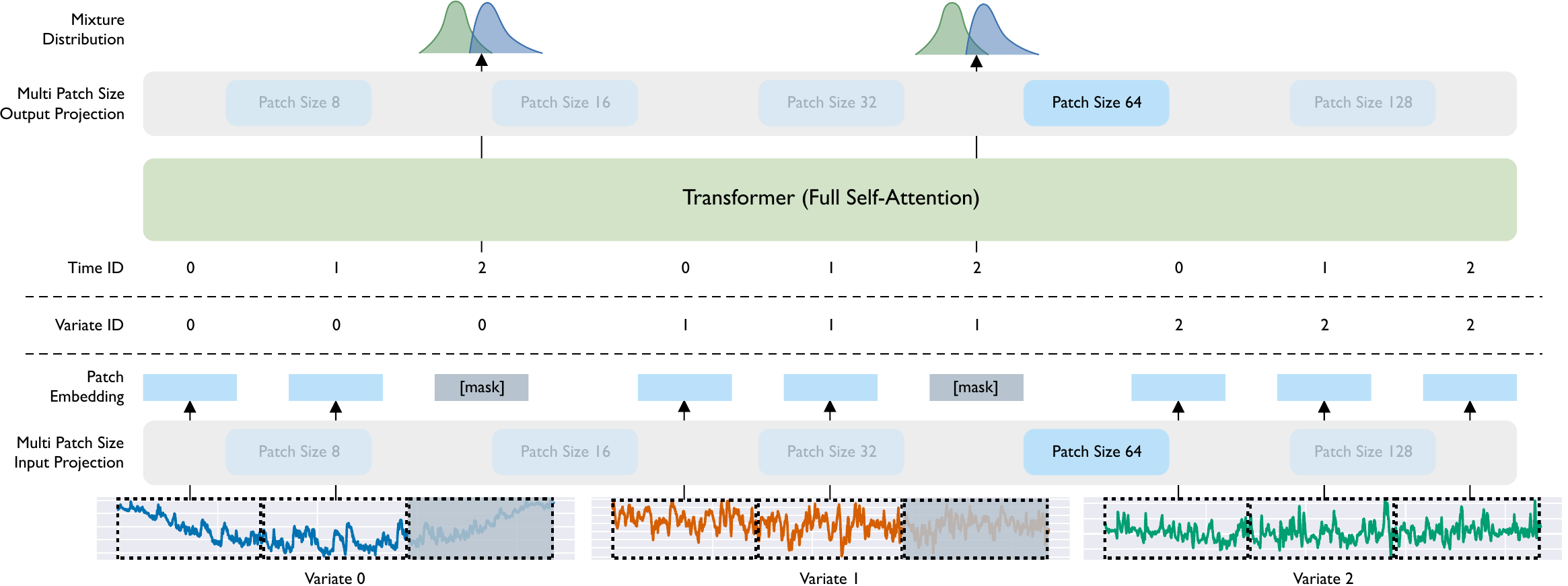}}
\caption{
Overall architecture of {\modelname}. Visualized is a 3-variate time series, where variates 0 and 1 are target variables (i.e. to be forecasted, and variate 2 is a dynamic covariate (values in forecast horizon known). Based on a patch size of 64, each variate is patchified into 3 tokens. The patch embeddings along with sequence and variate id are fed into the Transformer. The shaded patches represent the forecast horizon to be forecasted, whose corresponding output representations are mapped into the mixture distribution parameters.
}
\label{fig:architecture}
\end{center}
\end{figure*}

%% file: section/3_2_method.tex
\input{table/lotsa_summary}
\subsection{Unified Training}
\subsubsection{{\lotsa} Data}
Existing work has predominantly relied on three primary sources of data -- the Monash Time Series Forecasting Archive \citep{godahewa2021monash}, datasets provided by the GluonTS library \citep{alexander2020gluonts}, and datasets from the popular long sequence forecasting benchmark \citep{lai2018modeling,wu2021autoformer}.
While Monash and GluonTS comprise of datasets from diverse domains, they are constrained in size, with approximately \(1\mathrm{B}\) observations combined. In comparison, LLMs are trained on \textit{trillions} of tokens. \citet{das2023predct} builds a private dataset mainly based on Google Trends and Wiki pageviews, but lacks diversity in terms of the domains these time series originate from. \looseness=-1

The effectiveness of FMs heavily stem from large-scale pre-training data. Given that existing data sources fall short of supporting such a paradigm, attempting to train an LTM on them may result in misleading conclusions.
Thus, we tackle this issue head-on by building a large-scale archive of open time series datasets by collating publicly available sources of time series datasets. 
This effort aims to cover a broad spectrum of domains, consolidating datasets from diverse sources with varying formats. We design a unified storage format using Arrow \citep{richardson2023arrow} which is ready for deep learning pipelines.
The resulting collection, {\lotsa}, spans nine domains, with a total of \(27,646,462,733\) observations, with key statistics in \cref{tab:lotsa_summary_domain,tab:lotsa_summary_freq}, and in-depth details in \cref{app:lotsa}.

\subsubsection{Pre-training}
As introduced in \cref{eq:optim_problem}, our pre-training task is formulated to optimize the mixture distribution log-likelihood. 
The design of both the data distribution and task distribution are two critical aspects of the pre-training pipeline. This design imparts versatile capabilities to our LTM, enabling it to adapt to a range of downstream tasks. This flexibility stands in contrast to the prevailing deep forecasting paradigm, where models are typically specialized for specific datasets and settings.

\paragraph{Data Distribution}
The data distribution, \((\rmY, \rmZ) \sim p(\gD)\), defines how time series are sampled from the dataset. Trained on {\lotsa}, which is a dataset of datasets, we introduce the notion of sub-datasets, by decomposing the data distribution into a sub-dataset distribution, and a time series distribution conditioned on a sub-dataset, \(p(\gD) = p(\rmY, \rmZ | \rmD) p(\rmD)\).
Thus, we first sample a sub-dataset from \(p(\rmD)\), and given that sub-dataset, we sample a time series.
For \(K\) sub-datasets, where \(\mD_k\) represents the set of indices of time series belonging to sub-dataset \(k\),
the structure of 
\(p(\mY^{(i)}, \mZ^{(i)}|\mD_k) = \frac{T_i * \mathbb{1}_{\{i \in \mD_k\}}}{\sum_{j \in \mD_k} T_j}\), proportionate to the number of observations, is straightforward.

However, due to data imbalance across domains and frequency, we avoid sampling sub-datasets proportionately, and instead cap the contribution of each sub-dataset at \(\epsilon = 0.001\), before re-normalizing: \(p(\mD_k) = \frac{\omega_k}{\sum_{i=1}^K \omega_i}\), where \(\omega_k = \min(\frac{|\mD_k|}{\sum_{i}^K |\mD_i|}, \epsilon)\), and \(|\mD_k| = \sum_{i \in \mD_k} T_i\).

\paragraph{Task Distribution}
Different from the existing deep forecasting paradigm, we aim to train a model with forecasting capabilities over varying context and prediction lengths. Rather than defining a fixed context and prediction length, we sample from a task distribution, \((\rt, \rl, \rh) \sim p(\gT|\gD)\) which defines the lookback window and forecasting horizon, given a time series. 
In practice, rather than sampling \(t, l, h\), given a time series, we crop a uniformly sampled window, whose length is uniformly sampled from a range. This range is defined by a minimum sequence length per variate of \(2\), and a total maximum sequence length of \(512\). The window is then split into lookback and horizon segments, where the prediction length is uniformly sampled as a proportion (within the range \([0.15, 0.5]\)) of the window.
We further augment training by 
\(i)\) uniformly subsampling multivariate time series in the variate dimension,
and \(ii)\) constructing multivariate time series from sub-datasets with univariate time series, by randomly concatenating them. The number of variates is sampled from a beta-binomial distribution with parameters \(n=128, a=2, b=5\) which supports a maximum of \(128\) variates, with mean \(\approx 37\) for efficiency. \looseness=-1

\paragraph{Training}
\input{table/model_size}
We train {\modelname} in three sizes -- small, base, and large, with key parameter details listed in \cref{tab:model_size}.
The small model is trained for \(100,000\) steps, while base and large models are trained for \(1,000,000\) steps with a batch size of 256. 
For optimization, we use the AdamW optimizer with the following hyperparameters, \(\text{lr}=1\mathrm{e}\text{-}{3}, \text{weight\_decay}=1\mathrm{e}\text{-}{1}, \beta_1=0.9, \beta_2=0.98\). We also apply a learning rate scheduler with linear warmup for the first \(10,000\) steps, and cosine annealing thereafter. Models are trained on NVIDIA A100-40G GPUs with TF32 precision.
We implement sequence packing \citep{raffel2020t5} to avoid large amounts of padding due to the disparity of sequence lengths in the new setting with varying context, prediction, and variate lengths, thereby increasing the effective batch size.
\looseness=-1

%% file: table/lotsa_summary.tex
\begin{table*}[t]
    \caption{Key statistics of {\lotsa} by domain.}
    \label{tab:lotsa_summary_domain}
    \centering
    \vskip 0.05in
    \resizebox{\textwidth}{!}{
    \begin{tabular}{lccccccccc}
        \toprule
            &
            \textbf{Energy} &
            \textbf{Transport} &
            \textbf{Climate} &
            \textbf{CloudOps} &
            \textbf{Web} &
            \textbf{Sales} &
            \textbf{Nature} &
            \textbf{Econ/Fin} &
            \textbf{Healthcare} \\
            
        \midrule
            \textbf{\# Datasets} & 
            30 & 
            23 & 
            6 & 
            3 & 
            3 &
            6 & 
            5 & 
            23 & 
            6 \\
            
            \textbf{\# Obs.} & 
            16,358,600,896 & 
            4,900,453,419 & 
            4,188,011,890 & 
            1,518,268,292 &
            428,082,373 & 
            197,984,339 & 
            28,547,647 & 
            24,919,596 & 
            1,594,281 \\
            
            \textbf{\%} & 
            59.17\% & 
            17.73\% & 
            15.15\% & 
            5.49\% &
            1.55\% & 
            0.72\% & 
            0.09\% & 
            0.10\% & 
            0.01\% \\
        \bottomrule
    \end{tabular}
    }
    \vskip -0.1in
\end{table*}

\begin{table*}[t]
    \caption{Key statistics of {\lotsa} by frequency.}
    \label{tab:lotsa_summary_freq}
    \vskip 0.05in
    \centering
    \resizebox{\textwidth}{!}{
    \begin{tabular}{lcccccccc}
        \toprule
            &
            \multicolumn{1}{c}{\textbf{Yearly}} &
            \multicolumn{1}{c}{\textbf{Quarterly}} &
            \multicolumn{1}{c}{\textbf{Monthly}} &
            \multicolumn{1}{c}{\textbf{Weekly}} &
            \multicolumn{1}{c}{\textbf{Daily}} &
            \multicolumn{1}{c}{\textbf{(Multi) Hourly}} &
            \multicolumn{1}{c}{\textbf{(Multi) Minute-level}} &
            \multicolumn{1}{c}{\textbf{(Multi) Second-level}} \\
        \midrule
            \textbf{\# Datasets} & 4 & 5 & 10 & 7 & 21 & 31 & 25 & 2 \\
            \textbf{\# Obs.} & 873,297 & 2,312,027 & 11,040,648 & 18,481,871 & 709,017,118 & 19,875,993,973 & 7,013,949,430 & 14,794,369 \\
            \textbf{\%} & 0.003\% & 0.008\% & 0.040\% & 0.067\% & 2.565\% & 71.893\% & 25.370\% & 0.054\% \\
        \bottomrule
    \end{tabular}
    }
    \vskip -0.1in
\end{table*}

%% file: table/model_size.tex
\begin{table}[t]
    \caption{Details of {\modelname} model sizes.}
    \label{tab:model_size}%
    \vskip 0.05in
    \centering
    \resizebox{\columnwidth}{!}{
    \begin{tabular}{lcccccc}
        \toprule
            &
            \textbf{Layers} &
            \textbf{$\bm{d}$\textsubscript{model}} & 
            \textbf{$\bm{d}$\textsubscript{ff}} &
            \textbf{Heads} &
            \textbf{$\bm{d}$\textsubscript{kv}} &
            \textbf{Params}
            \\
        \midrule
            {\smallmodel} &
            6 &
            384 &
            1536 &
            6 &
            64 &
            \(14\mathrm{m}\)
            \\
            {\basemodel} &
            12 &
            768 &
            3072 &
            12 &
            64 &
            \(91\mathrm{m}\)
            \\
            {\largemodel} &
            24 &
            1024 &
            4096 &
            16 &
            64 &
            \(311\mathrm{m}\)
            \\
        \bottomrule
    \end{tabular}%
    }
\end{table}%

%% file: section/4_0_experiments.tex
\section{Experiments}

\input{section/4_1_experiments_in}
\input{section/4_2_experiments_out}
\input{section/4_3_experiments_ablation}
\input{section/4_4_experiments_analysis}

%% file: section/4_1_experiments_in.tex
\subsection{In-distribution Forecasting}
\input{figure/monash_summary}
We first perform an in-distribution evaluation using the Monash benchmark, which aim to measure generalization capability across diverse domains. Described in \cref{app:lotsa_monash}, {\lotsa} includes the Monash Time Series Forecasting Archive as a source of data. For a large portion of these datasets, we only include the train set, holding out the test set which we now use for in-distribution evaluation. 
In this evaluation, we consider a standard setting with a context length of 1000, and a patch size of 32 for all frequencies, except for quarterly data with a patch size of 8.
\cref{fig:monash_summary} summarizes the results based on the normalized mean absolute error (MAE), in comparison with the baselines presented in the Monash benchmark. It is worth noting that each baseline in the Monash benchmark is typically trained individually per dataset or per time series within a dataset. In contrast, {\modelname} stands out by being a single model evaluated across various datasets. Full results as well as a comparison with LLMTime \citep{gruver2023llmtime} can be found in \cref{app:results_monash}.

We observe that {\modelname} outperforms all baselines from the Monash benchmark regardless of model size, displaying the strong in-distribution and cross-domain capabilities arising from our unified training methodology. We highlight that each instance of {\modelname} is a \textbf{single} model evaluated across datasets, compared to baselines for which one model is trained per dataset. Further analysis on computational costs can be found in \cref{app:comp_cost}.

%% file: figure/monash_summary.tex
\begin{figure}[t]
    \centering
    \includegraphics[width=\columnwidth]{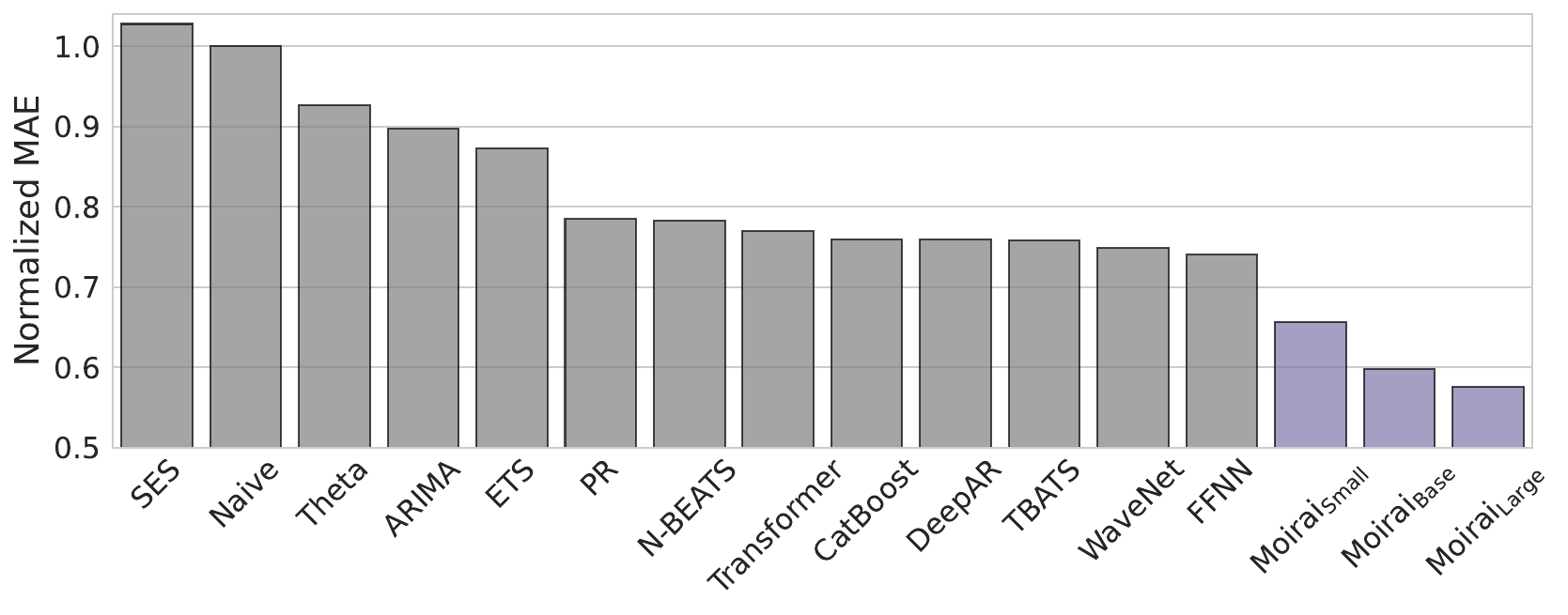}
\vskip -0.2 in
\caption{
Aggregate results of the Monash Time Series Forecasting Benchmark. The normalized MAE is reported, which normalizes the MAE of each dataset by the naive forecast's MAE, and aggregated by taking the geometric mean across datasets.
}
\label{fig:monash_summary}
\vskip -0.1in
\end{figure}

%% file: section/4_2_experiments_out.tex
\subsection{Out-of-distribution / Zero-shot Forecasting}
Next, we perform an out-of-distribution evaluation on \textbf{unseen target datasets}. Here, {\modelname} is a zero-shot forecaster compared with state-of-the-art full-shot baselines which have been trained on the individual target datasets.
While the ideal scenario would be to include other universal forecasters, this proves to be a challenging task. As a nascent field, most universal forecasters currently do not yet have open weights avaiable for evaluation.
Furthermore, the problem of comparing zero-shot methods is exacerbated by not having a standard held-out test split, making it challenging to collate a set of datasets which all the models have not been trained on.
Thus, we establish the strong zero-shot capabilities of {\modelname} by displaying competitive or stronger results compared with SOTA full-shot methods -- datasets used in the following have \textbf{not} been included in {\lotsa}.

\input{table/pf_summary}
\input{table/lsf_summary}

\paragraph{Probabilistic Forecasting}
We evaluate on six datasets across energy, transport, climate, and sales domains, following a rolling evaluation setup with stride equal to prediction length. Prediction lengths and number of rolling evaluations are defined for each dataset based on frequency.
We report the Continuous Ranked Probability Score (CRPS) and Mean Scaled Interval Score (MSIS) metrics (definitions in \cref{app:eval_details_pf}), comparing against four full-shot baselines -- DeepAR \citep{salinas2020deepar}, PatchTST \citep{nie2023patchtst}, and TiDE \citep{das2023tide} with Student's t-distribution prediction heads, and TFT based on quantile prediction \citep{lim2021tft}, all implemented with the GluonTS library \citep{alexander2020gluonts}, as well as simple baselines AutoARIMA \citep{garza2022statsforecast} and Seasonal Naive \citep{hyndman2018fpp}. For each dataset and baseline, we perform hyperparameter tuning on a validation CRPS, and report results averaged over five training runs with different seeds.
For {\modelname}, we perform inference time tuning, selecting context length from \(\{1000,2000,3000,4000,5000\}\) and patch sizes based on frequency, on the validation CRPS.
Full details of the evaluation setting can be found in \cref{app:eval_details_pf}.
\looseness=-1

\cref{tab:pf_summary} reports the CRPS and MSIS, with full results including deterministic metrics in \cref{app:results_pf}.
We observe that {\basemodel} and {\largemodel} consistently achieve strong zero-shot performance, obtaining either best or second best results for all datasets except Walmart and Istanbul Traffic. Even for these datasets, performance is still close to the best performance, despite being a single zero-shot model compared to baselines which have been tuned and trained on the train sets.
\looseness=-1

\paragraph{Long Sequence Forecasting}
We evaluate on a subset of the popular long sequence forecasting benchmark \citep{wu2021autoformer}, omitting datasets which have datasets from the same source present in our pre-training data and cannot be considered zero-shot. 
We report the Mean Squared Error (MSE) and MAE, comparing against six state-of-the-art baselines, iTransformer \citep{liu2023itransformer}, TimesNet \citep{wu2023timesnet}, PatchTST, Crossformer \citep{zhang2023crossformer}, TiDE, DLinear \citep{zeng2023dlinear}, SCINet \citep{liu2022scinet}, and FEDformer \citep{zhou2022fedformer}. Point forecasts are obtained from {\modelname} by taking the median from the samples of the predictive distribution. Tuning for {\modelname} was based on the average validation MSE across prediction lengths, further including the options between channel indepedent and channel mixing strategies \citep{nie2023patchtst} for the low dimension datasets (ETT and Weather).

\cref{tab:lsf_summary} reports the average performance across prediction lengths, with full results in \cref{app:results_lsf}.
We observe that {\modelname} achieves strong results compared to full-shot baselines. While {\basemodel} consistently achieves strong performance across datasets with either best or second-best performance, the large model is less consistent, with slightly weaker but competitive results. The relationship between performance and model size is tenuous in this setting, however, this does not constitute strong evidence against the potential of scaling, since these results are based on models trained on a fixed dataset size and settings. Rather, this calls for more comprehensive neural scaling laws \citep{kaplan2020scaling} for LTMs, to build a stronger understanding of their scaling behavior.
\looseness=-1

%% file: table/pf_summary.tex
\begin{table*}[ht]
  \centering
  \caption{
  Probabilistic forecasting results. Best results are highlighted in \textbf{bold}, and second best results are \underline{underlined}. Baseline results are aggregated over five training runs with different seeds, reporting the mean and standard deviation.
  }
  \label{tab:pf_summary}%
\resizebox{0.875\textwidth}{!}{
\begin{tabular}{lcccccccccc}
\toprule
&       & \multicolumn{3}{c}{\textbf{Zero-shot}} & \multicolumn{4}{c}{\textbf{Full-shot}} & \multicolumn{2}{c}{\textbf{Baseline}} \\
\cmidrule(lr){3-5} \cmidrule(lr){6-9} \cmidrule(lr){10-11}
      &       & \textbf{{\smallmodel}} & \textbf{{\basemodel}} & \textbf{{\largemodel}} & \textbf{PatchTST} & \textbf{TiDE} & \textbf{TFT} & \textbf{DeepAR} & \textbf{AutoARIMA} & \textbf{Seasonal Naive} \\
\midrule
\multirow{2}[2]{*}{Electricity} & \textbf{CRPS} & 0.072 & 0.055 & \underline{0.050} & 0.052{\scriptsize$\pm$}0.00 & \boldmath{}\textbf{0.048{\scriptsize$\pm$}0.00}\unboldmath{} & 0.050{\scriptsize$\pm$}0.00 & 0.065{\scriptsize$\pm$}0.01 & 0.327 & 0.070 \\
      & \textbf{MSIS} & 7.999 & 6.172 & 5.875 & \underline{5.744{\scriptsize$\pm$}0.12} & \boldmath{}\textbf{5.672{\scriptsize$\pm$}0.08}\unboldmath{} & 6.278{\scriptsize$\pm$}0.24 & 6.893{\scriptsize$\pm$}0.82 & 29.412 & 35.251 \\
\midrule
\multirow{2}[2]{*}{Solar} & \textbf{CRPS} & 0.471 & \underline{0.419} & \textbf{0.406} & 0.518{\scriptsize$\pm$}0.09 & 0.420{\scriptsize$\pm$}0.00 & 0.446{\scriptsize$\pm$}0.03 & 0.431{\scriptsize$\pm$}0.01 & 1.055 & 0.512 \\
      & \textbf{MSIS} & 8.425 & \underline{7.011} & \textbf{6.250} & 8.447{\scriptsize$\pm$}1.59 & 13.754{\scriptsize$\pm$}0.32 & 8.057{\scriptsize$\pm$}3.51 & 11.181{\scriptsize$\pm$}0.67 & 25.849 & 48.130 \\
\midrule
\multirow{2}[2]{*}{Walmart} & \textbf{CRPS} & 0.103 & 0.093 & 0.098 & \underline{0.082{\scriptsize$\pm$}0.01} & \boldmath{}\textbf{0.077{\scriptsize$\pm$}0.00}\unboldmath{} & 0.087{\scriptsize$\pm$}0.00 & 0.121{\scriptsize$\pm$}0.00 & 0.124 & 0.151 \\
      & \textbf{MSIS} & 9.371 & 8.421 & 8.520 & \boldmath{}\textbf{6.005{\scriptsize$\pm$}0.21}\unboldmath{} & \underline{6.258{\scriptsize$\pm$}0.12} & 8.718{\scriptsize$\pm$}0.10 & 12.502{\scriptsize$\pm$}0.03 & 9.888 & 49.458 \\
\midrule
\multirow{2}[2]{*}{Weather} & \textbf{CRPS} & 0.049 & \textbf{0.041} & 0.051 & 0.059{\scriptsize$\pm$}0.01 & 0.054{\scriptsize$\pm$}0.00 & \underline{0.043{\scriptsize$\pm$}0.00} & 0.132{\scriptsize$\pm$}0.11 & 0.252 & 0.068 \\
      & \textbf{MSIS} & 5.236 & \underline{5.136} & \textbf{4.962} & 7.759{\scriptsize$\pm$}0.49 & 8.095{\scriptsize$\pm$}1.74 & 7.791{\scriptsize$\pm$}0.44 & 21.651{\scriptsize$\pm$}17.34 & 19.805 & 31.293 \\
\midrule
\multirow{2}[2]{*}{Istanbul Traffic} & \textbf{CRPS} & 0.173 & 0.116 & 0.112 & 0.112{\scriptsize$\pm$}0.00 & 0.110{\scriptsize$\pm$}0.01 & \underline{0.110{\scriptsize$\pm$}0.01} & \boldmath{}\textbf{0.108{\scriptsize$\pm$}0.00}\unboldmath{} & 0.589 & 0.257 \\
      & \textbf{MSIS} & 5.937 & 4.461 & 4.277 & \boldmath{}\textbf{3.813{\scriptsize$\pm$}0.09}\unboldmath{} & 4.752{\scriptsize$\pm$}0.17 & \underline{4.057{\scriptsize$\pm$}0.44} & 4.094{\scriptsize$\pm$}0.31 & 16.317 & 45.473 \\
\midrule
\multirow{2}[2]{*}{Turkey Power} & \textbf{CRPS} & 0.048 & 0.040 & \textbf{0.036} & 0.054{\scriptsize$\pm$}0.01 & 0.046{\scriptsize$\pm$}0.01 & \underline{0.039{\scriptsize$\pm$}0.00} & 0.066{\scriptsize$\pm$}0.02 & 0.116 & 0.085 \\
      & \textbf{MSIS} & 7.127 & \underline{6.766} & \textbf{6.341} & 8.978{\scriptsize$\pm$}0.51 & 8.579{\scriptsize$\pm$}0.52 & 7.943{\scriptsize$\pm$}0.31 & 13.520{\scriptsize$\pm$}1.17 & 14.863 & 36.256 \\
\bottomrule
\end{tabular}%
}
\vskip -0.05in
\end{table*}%

%% file: table/lsf_summary.tex
\begin{table*}[ht]
  \centering
  \caption{
  Long sequence forecasting results. Results are averaged across prediction lengths \(\{96,192,336,720\}\).
  Best results are highlighted in \textbf{bold}, and second best results are \underline{underlined}.
  Full-shot results are obtained from \citet{liu2023itransformer}.
  }
  \label{tab:lsf_summary}%
  \resizebox{0.875\textwidth}{!}{
\begin{tabular}{lcccccccccccc}
\toprule
&       & \multicolumn{3}{c}{\textbf{Zero-shot}} & \multicolumn{8}{c}{\textbf{Full-shot}} \\
\cmidrule(lr){3-5} \cmidrule(lr){6-13}
&       & \textbf{{\smallmodel}} & \textbf{{\basemodel}} & \textbf{{\largemodel}} & \textbf{iTransformer} & \textbf{TimesNet} & \textbf{PatchTST} & \textbf{Crossformer} & \textbf{TiDE} & \textbf{DLinear} & \textbf{SCINet} & \textbf{FEDformer} \\
\midrule
\multirow{2}[2]{*}{ETTh1} & \textbf{MSE} & \textbf{0.400} & \underline{0.434} & 0.510 & 0.454 & 0.458 & 0.469 & 0.529 & 0.541 & 0.456 & 0.747 & 0.44 \\
      & \textbf{MAE} & \textbf{0.424} & \underline{0.438} & 0.469 & 0.448 & 0.450 & 0.455 & 0.522 & 0.507 & 0.452 & 0.647 & 0.46 \\
\midrule
\multirow{2}[2]{*}{ETTh2} & \textbf{MSE} & \textbf{0.341} & \underline{0.345} & 0.354 & 0.383 & 0.414 & 0.387 & 0.942 & 0.611 & 0.559 & 0.954 & 0.437 \\
      & \textbf{MAE} & \underline{0.379} & 0.382 & \textbf{0.376} & 0.407 & 0.497 & 0.407 & 0.684 & 0.550 & 0.515 & 0.723 & 0.449 \\
\midrule
\multirow{2}[2]{*}{ETTm1} & \textbf{MSE} & 0.448 & \textbf{0.381} & 0.390 & 0.407 & 0.400 & \underline{0.387} & 0.513 & 0.419 & 0.403 & 0.486 & 0.448 \\
      & \textbf{MAE} & 0.409 & \textbf{0.388} & \underline{0.389} & 0.410 & 0.406 & 0.400 & 0.495 & 0.419 & 0.407 & 0.481 & 0.452 \\
\midrule
\multirow{2}[2]{*}{ETTm2} & \textbf{MSE} & 0.300 & \textbf{0.272} & \underline{0.276} & 0.288 & 0.291 & 0.281 & 0.757 & 0.358 & 0.35  & 0.571 & 0.305 \\
      & \textbf{MAE} & 0.341 & \underline{0.321} & \textbf{0.320} & 0.332 & 0.333 & 0.326 & 0.611 & 0.404 & 0.401 & 0.537 & 0.349 \\
\midrule
\multirow{2}[2]{*}{Electricity} & \textbf{MSE} & 0.233 & 0.188 & \underline{0.188} & \textbf{0.178} & 0.193 & 0.216 & 0.244 & 0.252 & 0.212 & 0.268 & 0.214 \\
      & \textbf{MAE} & 0.320 & 0.274 & \underline{0.273} & \textbf{0.270} & 0.295 & 0.304 & 0.334 & 0.344 & 0.3   & 0.365 & 0.327 \\
\midrule
\multirow{2}[2]{*}{Weather} & \textbf{MSE} & \underline{0.242} & \textbf{0.238} & 0.259 & 0.258 & 0.259 & 0.259 & 0.259 & 0.271 & 0.265 & 0.292 & 0.309 \\
      & \textbf{MAE} & \underline{0.267} & \textbf{0.261} & 0.275 & 0.278 & 0.287 & 0.281 & 0.315 & 0.320 & 0.317 & 0.363 & 0.36 \\
\bottomrule
\end{tabular}%

    }
\end{table*}%

%% file: section/4_3_experiments_ablation.tex
\vspace{-0.1in}
\subsection{Ablation Study}
\label{subsec:experiments_ablation}

\input{table/ablation}
\input{figure/ablation_distribution}
\vspace{-0.05in}
\paragraph{Architecture} 
We perform a series of ablations in \cref{tab:ablation}, starting from the default {\smallmodel}.
Firstly, we ablate the multi patch size component, removing the constraints by allowing any frequency to have any patch size during training, and also simply fixing the patch size to 32. In both cases, we observe a deterioration in normalized MAE.
Removing Any-variate Attention and using additive learned embeddings (randomizing variate index during training to encourage permutation invariance) instead, leads to suboptimal results, showcasing the strength of Any-variate Attention.
We see similar deterioration when replacing the mixture distribution with a Student's t-distribution, and further visualize the necessity of flexible distributions for probabilistic forecasts in \cref{fig:ablation_distribution}. \looseness=-1

\vspace{-0.1in}
\paragraph{Training Methodology}
We study the impact of a large and diverse dataset by training {\smallmodel} only on the GluonTS and Monash datasets, observing that diversity of data is critical for cross-domain training even on in-distribution evaluation.
Finally, given the same batch size and training iterations, we show that packed training significantly boosts performance. This is because packing increases the effective batch size and increases the number of observations the model is trained on, given the same amount of compute.

%% file: table/ablation.tex
\begin{table}[t]
  \centering
    \vskip -0.1in
  \caption{
  Ablation study on Monash benchmark. The aggregated normalized MAE, similarly calculated as in \cref{fig:monash_summary} is reported.
  }
  \label{tab:ablation}%
  \resizebox{0.75\columnwidth}{!}{
\begin{tabular}{lc}
\toprule
      & \textbf{Normalized MAE} \\
\midrule
{\smallmodel} & \textbf{0.655} \\
\quad w/o patch size constraints & 0.720 \\
\quad w/o multi patch size & 1.156 \\
\quad w/o Any-variate Attention & 0.904 \\
\quad w/o mixture distribution & 0.740 \\
\quad w/o {\lotsa} & 0.809 \\
\quad w/o packing & 0.785 \\
\bottomrule
\end{tabular}%
  }
\end{table}%

%% file: figure/ablation_distribution.tex
\begin{figure}[t]
\begin{center}
    \begin{subfigure}[t]{\columnwidth}
        \centering
        \includegraphics[width=0.9\columnwidth]{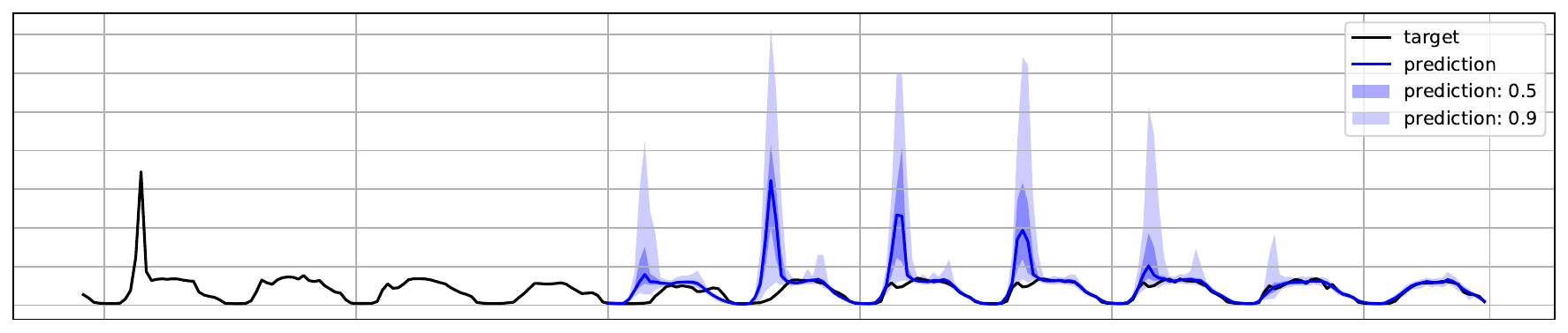}
        \caption{Mixture distribution.}
    \end{subfigure}
    \bigskip
    \begin{subfigure}[t]{\columnwidth}
        \centering
        \includegraphics[width=0.9\columnwidth]{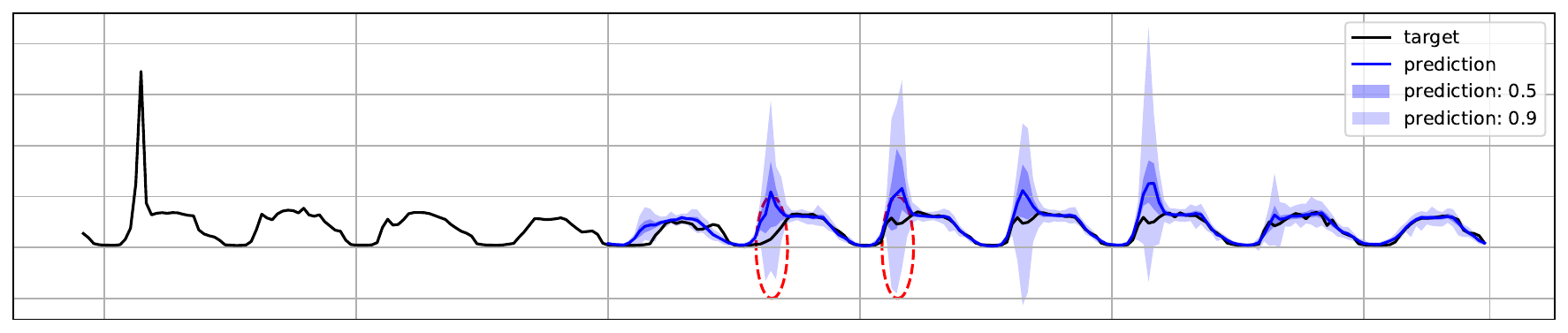}
        \caption{Student's t-distribution.}
    \end{subfigure}
\vskip -0.2in
\caption{
Visualization of probabilistic forecasts by two variants of {\smallmodel} on the Traffic Hourly dataset. Both models forecast peaks, however, the Student's t-distribution has a symmetric distribution, giving inappropriate prediction intervals for a peak, as highlighted in red.
}
\label{fig:ablation_distribution}
\end{center}
\vskip -0.1in
\end{figure}

%% file: section/4_4_experiments_analysis.tex
\subsection{Further Analysis}
\vspace{-0.05in}
\input{figure/context_length}
\paragraph{Context Length}
Our pre-training methodology varies context length defined by the task distribution. We verify that {\modelname} has the capability to take as input arbitrary context lengths by visualizing in \cref{fig:context_length} the relationship between performance and increasing context lengths over three datasets in the zero-shot setting. \citet{zeng2023dlinear,liu2023itransformer} previously observed that the desiderata of continuously improving performance with increasing context length is not present in conventional Transformer-based forecasters. Here, we observe that {\modelname} indeed achieves this desired property, in fact, capable of handling thousands of time steps. \looseness=-1

\input{figure/sequence_length_hist}
\vspace{-0.1in}
\paragraph{Packing}
Packing has long been applied in training LLMs and other Transformer-based models, but not for time series Transformers. While we can get away with inefficiencies when dealing with small-scale data, we start to suffer from longer training times as we scale towards the paradigm of FMs and LTMs. This is further exacerbated by our ``flattened'' setting which increases the disparity in sequence lengths. As evidenced in \cref{subsec:experiments_ablation}, keeping compute (batch size, iterations, etc.) constant, packing improves performance by \(16\%\). 
To understand why this is the case, we visualize sequence length distribution in \cref{fig:sequence_length_hist}. With a large portion of the data being shorter than the maximum sequence length, padding represents a whopping \(61.08\%\) of input tokens without packed training, and only \(0.38\%\) with our packed implementation (calculated over 1000 iterations). \looseness=-1

%% file: figure/context_length.tex
\begin{figure}[t]
\begin{center}
\centerline{\includegraphics[width=\columnwidth]{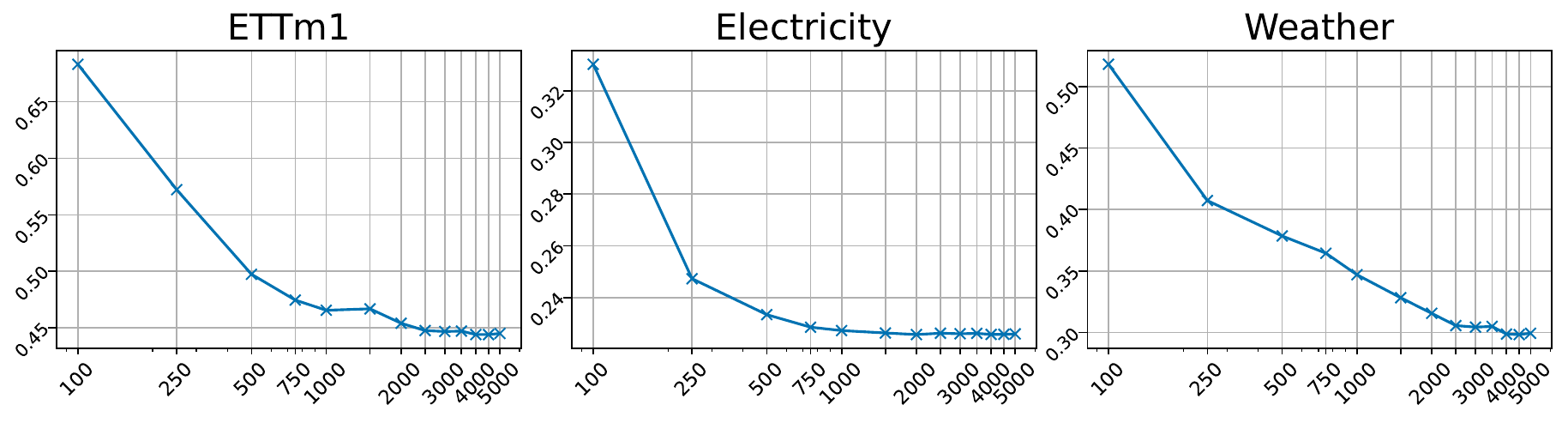}}
\vskip -0.1 in
\caption{
Plot of performance (MAE) against context length (x-axis in log scale) with prediction length 96 and patch size 32 on the validation set of the ETTm1, Electricity, and Weather datasets.
}
\label{fig:context_length}
\end{center}
\vskip -0.15in
\end{figure}

%% file: figure/sequence_length_hist.tex
\begin{figure}[t]
\begin{center}
\centerline{\includegraphics[width=\columnwidth]{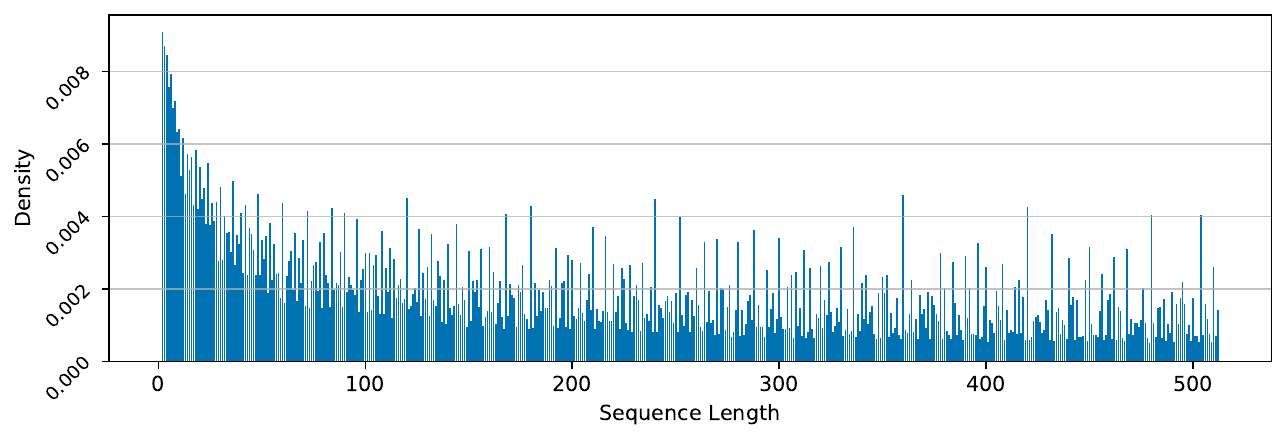}}
\vskip -0.2 in
\caption{
Histogram of sequence length when sampling data from {\lotsa} based on the proposed task distribution. Sequence length refers to the number of tokens after patching and flattening.
}
\label{fig:sequence_length_hist}
\end{center}
\vskip -0.2in
\end{figure}

%% file: section/5_conclusion.tex
\vspace{-0.05in}
\section{Conclusion}
In this work, we introduced {\modelname}, a masked encoder-based universal time series forecasting Transformer which alleviates the issues faced in the universal forecasting paradigm. We also introduce the {\lotsa}, the largest collection of open-data for pre-training time series forecasting models. {\modelname} is evaluated on the in-distribution and out-of-distribution settings, and is capable of probabilistic and long sequence forecasting. We show that as a zero-shot forecaster, {\modelname} achieves competitive or superior performance compared to full-shot models.

\vspace{-0.065in}
\paragraph{Limitations \& Future Work}
While {\modelname} achieves phenomenal in and out-of-distribution performance, this is just a first step in the universal forecasting paradigm. Due to resource constraints, little to no hyperparameter tuning was performed -- efficient tuning techniques such as \(\mu\)P \citep{yang2022tensor} can be applied.
In terms of architecture, our approach to tackling cross-frequency learning with a multi patch size mapping is somewhat heuristic, and future work should design a more flexible and elegant approach. Also, the current architecture has limited support for high-dimensional time series, and efficient methods for extending Transformer input length can alleviate this issue. The masked encoder structure also makes it amenable to exploration of a latent diffusion architecture \citep{feng2024latent}.
In terms of data, {\lotsa} can be further enhanced with greater diversity in terms of domain and frequency. 
Finally, incorporating multi-modality such as tabular or text inputs is an exciting new direction which universal forecasting has unlocked.

%% file: appendix/appendix.tex
\appendix
\onecolumn
\input{appendix/lotsa}
\input{appendix/architecture}
\newpage
\input{appendix/eval_details}
\input{appendix/results}
\input{appendix/forecast_viz}

%% file: appendix/lotsa.tex
\section{Large-scale Open Time Series Archive}
\label{app:lotsa}
{\lotsa} is a collection of time series datasets curated for pre-training of LTMs. In the following, we describe its constituent datasets and their respective sources, listing any pre-processing and data splitting performed. We further details on the key properties of these datasets, providing the domain, frequency, number of time series, number of target variates, number of past covariates (covariates whose values in the forecast horizon are unknown), and total number of observations in the dataset (defined as \(\sum_{i=1}^{N} T_i\) for a dataset with \(N\) time series).
Of note, if we consider number of observations to include the number of variates, i.e. \(\sum_{i=1}^N T_i * d_{y_i}\), {\lotsa} would have 231,082,956,489 (\(231\mathrm{B}\)) total observations.

\paragraph{BuildingsBench}
\input{table/lotsa_details/buildings_bench}

BuildingsBench \citep{emami2023buildingsbench} (\cref{tab:lotsa_buildings_bench}) provides a collection of datasets for residential and commercial building energy consumption. The BDG-2 datasets, Low Carbon London, SMART, IDEAL, Sceaux, and Borealis are real building energy consumption datasets from various sources.
The Electricity dataset \citep{trindade2015electricity} is also included in BuildingsBench but we omit it from {\lotsa} and use it for out-of-distribution evaluation instead.
They further release the Buildings-900K dataset a large-scale dataset of 900K simulated buildings. \citet{emami2023buildingsbench} introduce a pre-train/test split based on Public Use Microdata Area, but we use include both splits in {\lotsa} for pre-training.
No special pre-processing was applied to these datasets.
More information regarding these datasets can be found in \citet{emami2023buildingsbench}.
\looseness=-1

\paragraph{ClimateLearn}
\input{table/lotsa_details/climate_learn}

We include the ERA5 and CMIP6 datasets from the ClimateLearn library \citep{nguyen2023climatelearn} (\cref{tab:lotsa_climate_learn}). The ERA5 and CMIP6 datasets provide time series of various climate related variables such as temperature, and humidity and various pressure levels, based on a grid structure. We use the \(2.8125^{\circ}\) resolution which contains time series in a \(64 \times 128\) grid.

\paragraph{CloudOps TSF}
\input{table/lotsa_details/cloudops_tsf}

\citet{woo2023pushing} introduces three large-scale CloudOps time series datasets (\cref{tab:lotsa_cloudops_tsf}) measuring various variables such as CPU and memory utilization.
We follow their pre-train/test split and only include the pre-train time series in {\lotsa}, holding out the test set. \looseness=-1

\paragraph{GluonTS}
\input{table/lotsa_details/gluonts}

The GluonTS library \citep{alexander2020gluonts} provides many datasets popularly used in time series forecasting. We only include the datasets presented in \cref{tab:lotsa_gluonts} as we either hold out the other datasets, or are already included in the Monash repository.

\paragraph{LargeST}
\input{table/lotsa_details/largest}

LargeST \citep{liu2023largest} (\cref{tab:lotsa_largest}) collects the largest dataset from the California Department of Transportation Performance Measurement System (PeMS) \citep{chen2001pems} to date. The PeMS is a popular source of data for many traffic forecasting datasets such as PEMS03, PEMS04, PEMS07, PEMS08, and PEMS Bay, as well as the popular Traffic dataset from \citep{lai2018modeling}. Since we use such a large amount of dataset from the same source, we can no longer consider forecasting on any of these datasets as an out-of-distribution or zero-shot forecasting task anymore, and thus omit the Traffic dataset of the LSF benchmark from our evaluations.

\paragraph{LibCity}
\input{table/lotsa_details/lib_city}

LibCity \citep{wang2023libcity} (\cref{tab:lotsa_lib_city}) provides a collection urban spatio-temporal datasets. We drop the spatial aspect of these datsets and consider them as time series data.

\paragraph{Monash}
\label{app:lotsa_monash}
\input{table/lotsa_details/monash}

The Monash Time Series Forecasting Repository \citep{godahewa2021monash} (\cref{tab:lotsa_monash}) is a large collection of diverse time series datasets, the most popular source for building LTMs. We use Monash for in-distribution evaluation, and thus apart from the exceptions listed below, we only include the train region in {\lotsa}, by holding out the final forecast horizon as the test set. The final forecast horizon is defined for each dataset by \citep{godahewa2021monash}. 
For the following datasets, we include their entirety in {\lotsa} without a held-out test set for the following reasons:
\begin{itemize}
    \item London Smart Meters, Wind Farms, Wind Power, Solar Power, Oikolab Weather, Covid Mobility: Originally not included in the Monash benchmark.
    \item Extended Web Traffic, Kaggle Web Traffic Weekly: We include the extended version of Web Traffic which contains overlap with the original Web Traffic dataset.
    \item M1 Yearly, M1 Quarterly, M3 Yearly, M3 Quarterly, M4 Yearly, M4 Quarterly, Tourism Yearly: Some time series in these datasets are too short after train/test splits, thus we do not split them (setting a minimum of 16 time steps to achieve at least 2 patches).
\end{itemize}
We omit Electricity \citep{trindade2015electricity} and Solar \citep{lai2018modeling} datasets for out-of-distribution evaluation. Note that the ``Weather'' from Monash is a different dataset from that used in the out-of-distribution evaluations.

\paragraph{ProEnFo}
\input{table/lotsa_details/proenfo}

ProEnFo \citep{wang2023proenfo} (\cref{tab:lotsa_proenfo}) provides a range of datasets for load forecasting. Unlike BuildingsBench, ProEnFo provides various covariates such as temperature, humidity, and wind speed. We again omit Electricity \citep{trindade2015electricity} which is originally included in ProEnFo for out-of-distribution evaluations.

\paragraph{SubseasonalClimateUSA}
\input{table/lotsa_details/subseasonal}

The SubseasonalClimateUSA library \citep{mouatadid2023subseasonalclimateusa} (\cref{tab:lotsa_subseasonal}) provides climate time series data at a lower granularity (daily) for subseasonal level forecasting. We extract two datasets Subseasonal Precipitation which is the precipitation data from 1948 - 1978, and Subseasonal, which is precipitation and temperature data from 1979 - 2023. \looseness=-1

\paragraph{Others}
\input{table/lotsa_details/others}

Finally, detailed in \cref{tab:lotsa_others}, we further collect datasets from miscellaneous sources not provided by a library or collection. These datasets require more extensive pre-processing since they are not provided by a library, and are raw data instead. We take a standard approach of filtering out time series which are either too short, or have too many missing values. Fo each time series, we consider all variates to be targets, unless otherwise specified by the creators of the dataset (e.g. KDD Cup 2022 is a competition dataset, for which only the ``Patv'' variate is defined to be the target, with 9 other covariates).

%% file: table/lotsa_details/buildings_bench.tex
\begin{table}[ht]
  \centering
  \caption{Datasets and key properties from BuildingsBench.}
    \begin{tabular}{lcccccc}
    \toprule
    \textbf{Dataset} & \textbf{Domain} & \textbf{Frequency} & 
    \textbf{\# Time Series} & \textbf{\# Targets} & \textbf{\# Past Covariates} & 
    \textbf{\# Obs.} \\
    \midrule
    BDG-2 Panther & Energy & H     & 105   & 1     & 0     & 919,800 \\
    BDG-2 Fox & Energy & H     & 135   & 1     & 0     & 2,324,568 \\
    BDG-2 Rat & Energy & H     & 280   & 1     & 0     & 4,728,288 \\
    BDG-2 Bear & Energy & H     & 91    & 1     & 0     & 1,482,312 \\
    Low Carbon London & Energy & H     & 713   & 1     & 0     & 9,543,348 \\
    SMART & Energy & H     & 5     & 1     & 0     & 95,709 \\
    IDEAL & Energy & H     & 219   & 1     & 0     & 1,265,672 \\
    Sceaux & Energy & H     & 1     & 1     & 0     & 34,223 \\
    Borealis & Energy & H     & 15    & 1     & 0     & 83,269 \\
    Buildings900K & Energy & H     & 1,792,328 & 1     & 0     & 15,702,590,000 \\ 
    \bottomrule
    \end{tabular}%
  \label{tab:lotsa_buildings_bench}%
\end{table}%

%% file: table/lotsa_details/climate_learn.tex
\begin{table}[ht]
  \centering
  \caption{Datasets and key properties from ClimateLearn.}
    \begin{tabular}{lccccccc}
    \toprule
    \textbf{Dataset} & \textbf{Domain} & \textbf{Frequency} & 
    \textbf{\# Time Series} & \textbf{\# Targets} & \textbf{\# Past Covariates} & 
    \textbf{\# Obs.} \\
    \midrule
    CMIP6 & Climate & 6H    & 1,351,680 & 53    & 0     & 1,973,453,000 \\
    ERA5  & Climate & H     & 245,760 & 45    & 0     & 2,146,959,000 \\
    \bottomrule
    \end{tabular}%
  \label{tab:lotsa_climate_learn}%
\end{table}%

%% file: table/lotsa_details/cloudops_tsf.tex
\begin{table}[ht]
  \centering
  \caption{Datasets and key properties from CloudOps TSF}
  \resizebox{\textwidth}{!}{
    \begin{tabular}{lccccccc}
    \toprule
    \textbf{Dataset} & \textbf{Domain} & \textbf{Frequency} & 
    \textbf{\# Time Series} & \textbf{\# Targets} & \textbf{\# Past Covariates} & 
    \textbf{\# Obs.} \\
    \midrule
    Azure VM Traces 2017 & CloudOps & 5T    & 159,472 & 1     & 2     & 885,522,908 \\
    Borg Cluster Data 2011 & CloudOps & 5T    & 143,386 & 2     & 5     & 537,552,854 \\
    Alibaba Cluster Trace 2018 & CloudOps & 5T    & 58,409 & 2     & 6     & 95,192,530 \\
    \bottomrule
    \end{tabular}%
}
  \label{tab:lotsa_cloudops_tsf}%
\end{table}%

%% file: table/lotsa_details/gluonts.tex
\begin{table}[ht]
  \centering
  \caption{Datasets and key properties from the GluonTS library.}
    \begin{tabular}{lccccccc}
    \toprule
    \textbf{Dataset} & \textbf{Domain} & \textbf{Frequency} & 
    \textbf{\# Time Series} & \textbf{\# Targets} & \textbf{\# Past Covariates} & 
    \textbf{\# Obs.} \\
    \midrule
    Taxi  & Transport & 30T   & 67,984 & 1     & 0     & 54,999,060 \\
    Uber TLC Daily & Transport & D     & 262   & 1     & 0     & 47,087 \\
    Uber TLC Hourly & Transport & H     & 262   & 1     & 0     & 1,129,444 \\
    Wiki-Rolling & Web   & D     & 47,675 & 1     & 0     & 40,619,100 \\
    M5    & Sales & D     & 30,490 & 1     & 0     & 58,327,370 \\
    \bottomrule
    \end{tabular}%
  \label{tab:lotsa_gluonts}%
\end{table}%

%% file: table/lotsa_details/largest.tex
\begin{table}[ht]
  \centering
  \caption{Key properties of the LargeST Benchmark Dataset.}
    \begin{tabular}{lccccccc}
    \toprule
    \textbf{Dataset} & \textbf{Domain} & \textbf{Frequency} & 
    \textbf{\# Time Series} & \textbf{\# Targets} & \textbf{\# Past Covariates} & 
    \textbf{\# Obs.} \\
    \midrule
    LargeST & Transport & 5T    & 42,333 & 1     & 0     & 4,452,510,528 \\
    \bottomrule
    \end{tabular}%
  \label{tab:lotsa_largest}%
\end{table}%

%% file: table/lotsa_details/lib_city.tex
\begin{table}[ht]
  \centering
  \caption{Datasets and key properties from the LibCity library.}
    \begin{tabular}{lccccccc}
    \toprule
    \textbf{Dataset} & \textbf{Domain} & \textbf{Frequency} & 
    \textbf{\# Time Series} & \textbf{\# Targets} & \textbf{\# Past Covariates} & 
    \textbf{\# Obs.} \\
    \midrule
    PEMS03 & Transport & 5T    & 358   & 1     & 0     & 9,382,464 \\
    PEMS04 & Transport & 5T    & 307   & 3     & 0     & 5,216,544 \\
    PEMS07 & Transport & 5T    & 883   & 1     & 0     & 24,921,792 \\
    PEMS08 & Transport & 5T    & 170   & 3     & 0     & 3,035,520 \\
    PEMS Bay & Transport & 5T    & 325   & 1     & 0     & 16,937,700 \\
    Los-Loop & Transport & 5T    & 207   & 1     & 0     & 7,094,304 \\
    Loop Seattle & Transport & 5T    & 323   & 1     & 0     & 33,953,760 \\
    SZ-Taxi & Transport & 15T   & 156   & 1     & 0     & 464,256 \\
    Beijing Subway & Transport & 30T   & 276   & 2     & 11    & 248,400 \\
    SHMetro & Transport & 15T   & 288   & 2     & 0     & 1,934,208 \\
    HZMetro & Transport & 15T   & 80    & 2     & 0     & 146,000 \\
    Rotterdam & Transport & 2T    & 208   & 1     & 0     & 4,813,536 \\
    Q-Traffic & Transport & 15T   & 45,148 & 1     & 0     & 264,386,688 \\
    \bottomrule
    \end{tabular}%
  \label{tab:lotsa_lib_city}%
\end{table}%

%% file: table/lotsa_details/monash.tex
\begin{table}[htp]
  \centering
  \caption{Datasets and key properties from the Monash Time Series Forecasting Repository.}
  \resizebox{\columnwidth}{!}{
    \begin{tabular}{lccccccc}
    \toprule
    \textbf{Dataset} & \textbf{Domain} & \textbf{Frequency} & 
    \textbf{\# Time Series} & \textbf{\# Targets} & \textbf{\# Past Covariates} & 
    \textbf{\# Obs.} \\
    \midrule
    London Smart Meters & Energy & 30T   & 5,520 & 1     & 0     & 166,238,880 \\
    Wind Farms & Energy & T     & 337   & 1     & 0     & 172,165,370 \\
    Wind Power & Energy & 4S    & 1     & 1     & 0     & 7,397,147 \\
    Solar Power & Energy & 4S    & 1     & 1     & 0     & 7,397,222 \\
    Oikolab Weather & Climate & H     & 8     & 1     & 0     & 800,456 \\
    Elecdemand & Energy & 30T   & 1     & 1     & 0     & 17,520 \\
    Covid Mobility & Transport & D     & 362   & 1     & 0     & 148,602 \\
    Kaggle Web Traffic Weekly & Web   & W     & 145,063 & 1     & 0     & 16,537,182 \\
    Extended Web Traffic & Web   & D     & 145,063 & 1     & 0     & 370,926,091 \\
    M1 Yearly & Econ/Fin & Y     & 106   & 1     & 0     & 3,136 \\
    M1 Quarterly & Econ/Fin & Q     & 198   & 1     & 0     & 9,854 \\
    M1 Monthly & Econ/Fin & M     & 617   & 1     & 0     & 44,892 \\
    M3 Yearly & Econ/Fin & Y     & 645   & 1     & 0     & 18,319 \\
    M3 Quarterly & Econ/Fin & Q     & 756   & 1     & 0     & 37,004 \\
    M3 Monthly & Econ/Fin & M     & 1,428 & 1     & 0     & 141,858 \\
    M3 Other & Econ/Fin & Q     & 174   & 1     & 0     & 11,933 \\
    M4 Yearly & Econ/Fin & Y     & 22,739 & 1     & 0     & 840,644 \\
    M4 Quarterly & Econ/Fin & Q     & 24,000 & 1     & 0     & 2,214,108 \\
    M4 Monthly & Econ/Fin & M     & 48,000 & 1     & 0     & 10,382,411 \\
    M4 Weekly & Econ/Fin & W     & 359   & 1     & 0     & 366,912 \\
    M4 Hourly & Econ/Fin & H     & 414   & 1     & 0     & 353,500 \\
    M4 Daily & Econ/Fin & D     & 4,227 & 1     & 0     & 9,964,658 \\
    NN5 Daily & Econ/Fin & D     & 111   & 1     & 0     & 81,585 \\
    NN5 Weekly & Econ/Fin & W     & 111   & 1     & 0     & 11,655 \\
    Tourism Yearly & Econ/Fin & Y     & 419   & 1     & 0     & 11,198 \\
    Tourism Quarterly & Econ/Fin & Q     & 427   & 1     & 0     & 39,128 \\
    Tourism Monthly & Econ/Fin & M     & 366   & 1     & 0     & 100,496 \\
    CIF 2016 & Econ/Fin & M     & 72   & 1     & 0     & 6,334 \\
    Traffic Weekly & Transport & W     & 862   & 1     & 0     & 82,752 \\
    Traffic Hourly & Transport & H     & 862   & 1     & 0     & 14,978,112 \\
    Australian Electricity Demand & Energy & 30T   & 5     & 1     & 0     & 1,153,584 \\
    Rideshare & Transport & H     & 2,304 & 1     & 0     & 859,392 \\
    Saugeen & Nature & D     & 1     & 1     & 0     & 23,711 \\
    Sunspot & Nature & D     & 1     & 1     & 0     & 73,894 \\
    Temperature Rain & Nature & D     & 32,072 & 1     & 0     & 22,290,040 \\
    Vehicle Trips & Transport & D     & 329   & 1     & 0     & 32,512 \\
    Weather & Climate & D     & 3,010 & 1     & 0     & 42,941,700 \\
    Car Parts & Sales & M     & 2,674 & 1     & 0     & 104,286 \\
    FRED MD & Econ/Fin & M     & 107   & 1     & 0     & 76,612 \\
    Pedestrian Counts & Transport & H     & 66    & 1     & 0     & 3,130,762 \\
    Hospital & Healthcare & M     & 767   & 1     & 0     & 55,224 \\
    COVID Deaths & Healthcare & D     & 266   & 1     & 0     & 48,412 \\
    KDD Cup 2018 & Energy & H     & 270   & 1     & 0     & 2,897,004 \\
    Bitcoin & Econ/Fin & D     & 18    & 1     & 0     & 74,824 \\
    US Births & Healthcare & D     & 1     & 1     & 0     & 7,275 \\
    \bottomrule
    \end{tabular}%
    }
  \label{tab:lotsa_monash}%
\end{table}%

%% file: table/lotsa_details/proenfo.tex
\begin{table}[htp]
  \centering
  \caption{Datasets and key properties from the ProEnFo library.}
    \begin{tabular}{lccccccc}
    \toprule
    \textbf{Dataset} & \textbf{Domain} & \textbf{Frequency} & 
    \textbf{\# Time Series} & \textbf{\# Targets} & \textbf{\# Past Covariates} & 
    \textbf{\# Obs.} \\
    \midrule
    Covid19 Energy & Energy & H     & 1     & 1     & 6     & 31,912 \\
    GEF12 & Energy & H     & 20    & 1     & 1     & 788,280 \\
    GEF14 & Energy & H     & 1     & 1     & 1     & 17,520 \\
    GEF17 & Energy & H     & 8     & 1     & 1     & 140,352 \\
    PDB   & Energy & H     & 1     & 1     & 1     & 17,520 \\
    Spanish & Energy & H     & 1     & 1     & 1     & 35,064 \\
    BDG-2 Hog & Energy & H     & 24    & 1     & 5     & 421,056 \\
    BDG-2 Bull & Energy & H     & 41    & 1     & 3     & 719,304 \\
    BDG-2 Cockatoo & Energy & H     & 1     & 1     & 5     & 17,544 \\
    ELF   & Energy & H     & 1     & 1     & 0     & 21,792 \\
    \bottomrule
    \end{tabular}%
  \label{tab:lotsa_proenfo}%
\end{table}%

%% file: table/lotsa_details/subseasonal.tex
\begin{table}[htp]
  \centering
  \caption{Datasets and key properties from the SubseasonalClimateUSA library.}
    \begin{tabular}{lccccccc}
    \toprule
    \textbf{Dataset} & \textbf{Domain} & \textbf{Frequency} & 
    \textbf{\# Time Series} & \textbf{\# Targets} & \textbf{\# Past Covariates} & 
    \textbf{\# Obs.} \\
    \midrule
    Subseasonal & Climate & D     & 862   & 4     & 0     & 14,097,148 \\
    Subseasonal Precipitation & Climate & D     & 862   & 1     & 0     & 9,760,426 \\
    \bottomrule
    \end{tabular}%
  \label{tab:lotsa_subseasonal}%
\end{table}%

%% file: table/lotsa_details/others.tex
\begin{table}[htp]
  \centering
  \caption{Datasets and key properties from other miscellaneous sources.}
  \resizebox{\columnwidth}{!}{
    \begin{tabular}{lccccccc}
    \toprule
    \textbf{Dataset} & \textbf{Source} & \textbf{Domain} & \textbf{Frequency} & 
    \textbf{\# Time Series} & \textbf{\# Targets} & \textbf{\# Past Covariates} & 
    \textbf{\# Obs.} \\
    \midrule
    KDD Cup 2022 &    \citet{zhou2022kddcup2022}   & Energy & 10T   & 134   & 1     & 9     & 4,727,519 \\
    GoDaddy & Kaggle & Econ/Fin & M     & 3,135 & 2     & 0     & 128,535 \\
    Favorita Sales & Kaggle & Sales & D     & 111,840 & 1     & 0     & 139,179,538 \\
    Favorita Transactions & Kaggle & Sales & D     & 54    & 1     & 0     & 84,408 \\
    Restaurant & Kaggle & Sales & D     & 216   & 1     & 0     & 76,573 \\
    Hierarchical Sales &    \citet{mancuso2021hierarchical}   & Sales & D     & 118   & 1     & 0     & 212,164 \\
    China Air Quality &   \citet{zheng2015chinaair}    & Nature & H     & 437   & 6     & 0     & 5,739,234 \\
    Beijing Air Quality & \citet{chen2019beijingair}   & Nature & H     & 12    & 11    & 0     & 420,768 \\
    Residential Load Power &   \citet{bergmeir2023residential}    & Energy & T     & 271   & 3     & 0     & 145,994,559 \\
    Residential PV Power &   \citet{bergmeir2023residential}    & Energy & T     & 233   & 3     & 0     & 125,338,950 \\
    CDC Fluview ILINet & \citet{cdc}   & Healthcare & W     & 75    & 5     & 0     & 63,903 \\
    CDC Fluview WHO NREVSS & \citet{cdc}   & Healthcare & W     & 74    & 4     & 0     & 41,760 \\
    Project Tycho & \citet{van2018tycho} & Healthcare & W     & 1,258 & 1     & 0     & 1,377,707 \\
    \bottomrule
    \end{tabular}%
    }
  \label{tab:lotsa_others}%
\end{table}%

%% file: appendix/architecture.tex
\newpage
\section{{\modelname} Architecture Details}
\label{app:architecture_details}

\subsection{Multi Patch Size Projection Layers}
\label{app:architecture_details_patch_size}
Each multi patch size projection is a simple Linear layer, for input projections, mapping patch size to hidden state, and for output projections, mapping hidden state to distribution parameters. In practice, we pre-define the frequency to patch size mapping heuristically, selecting smaller patch sizes for low frequency data and larger patch sizes for high frequency data as follows:
\begin{itemize}
    \item Yearly, Quarterly: 8
    \item Monthly: 8, 16, 32
    \item Weekly, Daily: 16, 32
    \item Hourly: 32, 64
    \item Minute-level: 32, 64, 128
    \item Second-level: 64, 128
\end{itemize}
Note that we only learn one Linear layer per patch size, and share them across frequencies if there is overlap. This means that we learn five input projection layers and five output projection layers.

\subsection{Mixture Distribution}
\label{app:architecture_details_mixture}
As described in \citet{salinas2020deepar}, our model predicts the parameters of a probability distribution, in this case, a mixture distribution. We apply a softmax layer to the parameters associated to the mixture weights, constraining them to the probability simplex. The mixture components are as described.

\paragraph{Student's t-distribution}
A random variable \(\rx\) following the Student's t-distribution has p.d.f.:
\[p(x; \nu, \mu, \tau) = \frac{\Gamma(\frac{\nu + 1}{2})}{\Gamma(\frac{\nu}{2})\sqrt{\pi\nu}\tau} \bigg(1 + \frac{1}{\nu} \big (\frac{x-\mu}{\tau} \big)^2 \bigg)^{-(\nu+1)/2}\]
with parameters \(\nu > 0, \mu \in \R, \tau > 0\), the degrees-of-freedom (df), location, and scale parameters respectively, and \(\Gamma\) is the gamma function.
We predict the df, location, and scale parameters, and apply a softplus function for the positivity constraint. We further lower bound the df parameter to \(2\), since variance is undefined otherwise.

\paragraph{Log-normal distribution} A random variable \(\rx\) which follows a log-normal distribution has p.d.f.:
\[p(x; \mu, \sigma) = \frac{1}{x\sigma\sqrt{2\pi}} \exp \bigg(-\frac{(\ln x - \mu)^2}{2 \sigma^2} \bigg)\]
with parameters \(\mu \in \R, \sigma > 0\). We predict both parameters, applying softplus function for positivity.

\paragraph{Negative binomial distribution}
Following \citet{awasthi2022mle}, we implement a continuous extension of the negative binomial distribution. A random variable \(\rx\) following such a distribution has p.d.f.:
\[p(x; r, p) \propto \frac{\Gamma(x+r)}{\Gamma(x+1)\Gamma(r)} (1 - p)^r p^x\]
given parameters \(r > 0\) and \(p \in [0, 1]\), and \(\Gamma\) is the gamma function. We predict both parameters, applying the softplus function for positivity, and sigmoid function to constrain to a probability.

\paragraph{Low variance normal distribution}
A random variable \(\rx\) following a normal distribution has p.d.f.:
\[p(x; \mu, \sigma) = \frac{1}{\sigma\sqrt{2\pi}} \exp \bigg( -\frac{(x - \mu)^2}{2\sigma^2} \bigg)\]
where \(\mu \in \R, \sigma > 0\). For a low variance normal distribution, we only predict \(\mu\), and fix \(\sigma\) to a small number, in this case we fix \(\sigma = 1\mathrm{e}\text{-}{3}\).

\subsection{Discussion on ``Flexible Distribution''}
\label{app:flexible_distribution}
\cref{tab:model_comparison} categorizes various pre-trained forecasting models with the notion of a ``flexible distribution'' -- this is largely a qualitative categorization rather than a quantitative one. As of writing, only 3 other models considered probabilistic forecasting -- Lag-llama, TimeGPT, and LLMTime. The other models only considered point forecasts, and thus the concept of "flexible distribution" does not apply to them. The following are specific reasons on why we made the categorization for the 3 models which can handle probabilistic forecasting:
\begin{itemize}
    \item Lag-llama uses a Student-T distribution which is only able to model symmetric distributions. This is an inflexible distribution which is unable to model asymmetric distributions, as demonstrated in \cref{fig:ablation_distribution} of our paper. They also raise this issue in their paper (Section 4.3), citing the use of more expressive distribution heads such as normalizing flows and copulas in future work.
    \item TimeGPT uses conformal prediction to construct prediction intervals. We refer to a tweet\footnote{\url{https://twitter.com/nixtlainc/status/1694466983927153131}} from the creators, which claim: "Some prediction intervals don’t account for domain constraints. A few users highlighted intervals suggesting negative values for time series that only take positive values." Thus, we consider it to be inflexible.
    \item LLMTime uses a categorical distribution. In their paper (paragraph titled "Language models as flexible distributions" in Section 3), they demonstrated that this approach is a flexible distribution which can approximate many different kinds of continuous distributions.
\end{itemize}

%% file: appendix/eval_details.tex
\section{Probabilistic Forecasting}
\label{app:eval_details_pf}

\subsection{Evaluation Metrics}

\paragraph{Continuous Ranked Probability Score}
The CRPS \citep{gneiting2007strictly} is a probabilistic forecasting evaluation metric, given a predicted distribution with c.d.f. \(F\) and ground truth \(y\), it is defined as: 
\begin{align*}
    \textrm{CRPS} & = \int_0^1 2 \Lambda_{\alpha}(F^{-1}(\alpha), y) d\alpha \\
    \Lambda_{\alpha}(q, y) & = (\alpha - \1_\mathrm{y < q})(y - q),
\end{align*}
where \(\Lambda_{\alpha}\) is the \(\alpha\)-quantile loss, also known as the pinball loss at quantile level \(\alpha\).

In practice, the CRPS is intractable or computationally expensive to compute, and we also want to compute a normalized metric, thus we compute a normalized discrete approximation, the mean weighted sum quantile loss \citep{park2022quantile}, defined as the average of \(K\) quantiles:
\begin{align*}
    \textrm{CRPS} & \approx \frac{1}{K} \sum_{k=1}^K \textrm{wQL}[\alpha_k] \\
    \textrm{wQL}[\alpha] & = 2 \frac{\sum_{t} \Lambda_{\alpha}(\hat{q}_{t}(\alpha), y_t)}{\sum_{t} |y_{t}|},
\end{align*}
where \(\hat{q}_t(\alpha)\) is the predicted \(\alpha\)-quantile at time step \(t\). We take \(K = 9, \alpha_1 = 0.1, \alpha_2 = 0.2, \ldots, \alpha_9 = 0.9\) in practice.

\paragraph{Mean Scaled Interval Score}
The MSIS is a metric to evaluate uncertainty around point forecasts, introduced in the M4 Competition \citep{makridakis2020m4}. Given an upper bound prediction, \(U_t\), and lower bound prediction \(L_t\), the MSIS is defined as:
\[
\textrm{MSIS} = \frac{1}{h} \frac{\sum_{t=1}^h (U_t - L_t) + \frac{2}{a}(L_t - Y_t) \mathbb{1}_{\{Y_t < L_t\}} + \frac{2}{a}(Y_t - U_t) \mathbb{1}_{\{Y_t > U_t\}}}{\frac{1}{n-m}\sum_{t=m+1}^n |Y_t - Y_{t-m}|}
\]
where \(a = 0.05\) is the significance level for a 95\% prediction interval, over a forecast horizon of length \(h\), and \(m\) is the seasonal factor.

\subsection{Evaluation Setup}
\input{table/pf_datasets}
We perform evaluation in a non-overlapping rolling window fashion, i.e. stride is equal to prediction length. The test set is defined as the last \(h * r\) time steps where \(h\) is the prediction length of the forecast horizon, and \(r\) is the number of rolling evaluation windows. We take the validation set to be the last forecast horizon before the test set, and the train set to be everything before that. From \cref{tab:pf_datasets}, our evaluation spans four domains, from minute-level to weekly frequencies. Prediction length and rolling evaluations are defined for each dataset based on frequency, making day ahead predictions for sub-hourly frequencies for seven days, and eight week ahead predictions for 32 weeks for weekly frequency.

\subsection{Baselines}

\input{table/pf_baseline_hyperparam}
For the four deep learning baselines, DeepAR \citep{salinas2020deepar}, PatchTST \citep{nie2023patchtst}, TiDE \citep{das2023tide}, and TFT \citep{lim2021tft}, we perform hyperparameter tuning based on the values presented in \cref{tab:pf_baseline_hyperparam}, and also tune learning rate \([1\mathrm{e}\textrm{-}{6}, 1\mathrm{e}\textrm{-}{3}]\) in log scale, and the context length as \(l = m * h\), where \(m\) is tuned in the range \([2, 20]\), and \(h\) is the prediction length. We perform random search through these values over 15 training runs, and report results on 5 independent training runs based on the best validation CRPS.

%% file: table/pf_datasets.tex
\begin{table}[H]
  \centering
  \caption{Summary of datasets used in the out-of-distribution probabilistic forecasting evaluation setting.}
  \resizebox{0.75\columnwidth}{!}{
    \begin{tabular}{lcccc}
    \toprule
    \textbf{Dataset} & \textbf{Domain} & \textbf{Frequency} & \textbf{Prediction Length} & \textbf{Rolling Evaluations} \\
    \midrule
    Electricity \citep{trindade2015electricity} & Energy & H     & 24    & 7 \\
    Solar \citep{lai2018modeling} & Energy & H     & 24    & 7 \\
    Walmart \citep{walmart2014sales} & Sales & W     & 8     & 4 \\
    Weather & Climate & 10T   & 144   & 7 \\
    Istanbul Traffic \tablefootnote{https://www.kaggle.com/datasets/leonardo00/istanbul-traffic-index} & Transport & H     & 24    & 7 \\
    Turkey Power \tablefootnote{https://www.kaggle.com/datasets/dharanikra/electrical-power-demand-in-turkey} & Energy & H     & 24    & 7 \\
    \bottomrule
    \end{tabular}%
  }
  \label{tab:pf_datasets}%
\end{table}%

%% file: table/pf_baseline_hyperparam.tex
\begin{table}[H]
  \centering
  \caption{Hyperparameter search values for probabilistic forecasting evaluation baselines.}
  \resizebox{0.5\columnwidth}{!}{
    \begin{tabular}{lcc}
    \toprule
          & \textbf{Hyperparameter} & \textbf{Values} \\
    \midrule
    PatchTST & d\_model & \(\{64, 128, 256\}\) \\
          & num\_encoder\_layers & \([2,6]\) \\
    \midrule
    DeepAR & hidden\_size & \(\{64, 128, 256\}\) \\
          & num\_layers & \([2,6]\) \\
    \midrule
    TFT & hidden\_dim & \(\{64, 128, 256\}\) \\
    \midrule
    TiDE & hidden\_dim & \(\{64, 128, 256\}\) \\
          & num\_encoder\_layers = num\_decoder\_layers & \([2,6]\) \\
    \bottomrule
    \end{tabular}%
  }
  \label{tab:pf_baseline_hyperparam}%
\end{table}%

%% file: appendix/results.tex
\newpage
\section{Full Experimental Results}
\label{app:results}

\subsection{In-distribution Forecasting: Monash Time Series Forecasting Benchmark}
\label{app:results_monash}
\input{figure/monash_with_llmtime}
\input{table/monash_full}

We include the full breakdown of results for the Monash benchmark in \cref{tab:monash_full}, including two versions of LLMTime: GPT3.5 (our reproduction), and LLaMA2 (results from \citet{gruver2023llmtime}). GPT3.5 is our reproduction by running their code\footnote{https://github.com/ngruver/llmtime} on the full dataset, using GPT3.5-Turbo-Instruct since text-davinci-003 has been deprecated.
LLaMA2 only has results for a subset of datasets in \cref{tab:monash_full}, thus in \cref{fig:monash_with_llmtime}, we present two aggregated results, one aggregated on the full \cref{tab:monash_full}, and one on the subset with results available for LLaMA2.
As observed, {\modelname} retains the top rankings for with the base and large models in all settings.

\subsection{Out-of-distribution Forecasting: Probabilistic Forecasting}
\label{app:results_pf}
\input{table/pf_full}

\cref{tab:pf_full} provides the full results of the probabilistic forecasting experiments with additional point forecasting metrics, including the symmetric mean absolute percentage error (sMAPE) \citep{hyndman2014errors}, mean absolute scaled error (MASE) \citep{hyndman2006another}, normalized deviation (ND), and normalized root mean squared error (NRMSE) \citep{yu2016temporal}.

\subsection{Out-of-distribution Forecasting: Long Sequence Forecasting}
\label{app:results_lsf}
\input{table/lsf_full}

\cref{tab:lsf_full} provides the full breakdown of results for the long sequence forecasting experiments, listing results for each prediction length.

\newpage
\subsection{Computation Costs}
\label{app:comp_cost}
\input{table/comp_cost}

We perform an analysis on the computation cost of {\modelname} compared to other deep learning based models, while varying the context and prediction lengths. Overall, given the same model size and setting, the cost of inference compared to other deep learning models would be similar.
From an architecture perspective, {\modelname} has the following benefits:
\begin{itemize}
    \item Patch based inputs: This decreases the computation cost significantly by reducing the number of input tokens.
    \item Masked encoder architecture: Unlike decoder-only Transformers, the masked encoder architecture can make multi step predictions in a single forward pass. For decoder-only Transformers and RNNs, they need to autoregressively make predictions, making multiple forward passes for a multi step forecast. For long horizons, this can be quite costly.
\end{itemize}
Furthermore, compared to standard baselines, {\modelname} performs zero-shot forecasting. The standard baseline approach has to be trained (multiple times with hyperparameter tuning) for each dataset, leading to increased costs. As {\modelname} continues to be utilized on new datasets, the pre-training costs are amortized and only becomes cheaper, while standard approaches need to be trained over and over again on new datasets.
We note that while {\modelname} indeed incurs increased costs due to model size, inference is still highly competitive, taking under 1 second to construct forecasts even with extremely long context/prediction lengths.

%% file: figure/monash_with_llmtime.tex
\begin{figure}[htp]
    \centering
    \begin{subfigure}[t]{0.5\textwidth}
        \centering
        \includegraphics[width=\columnwidth]{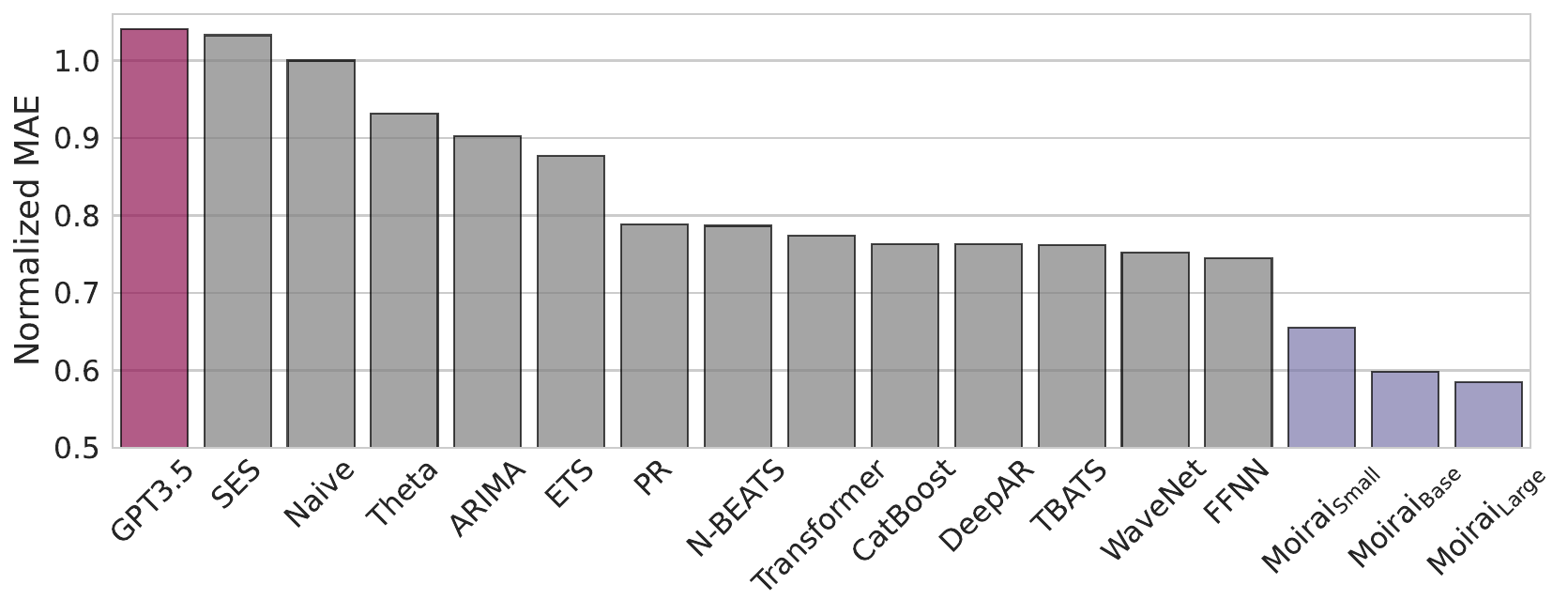}
        \caption{Results aggregated over full all datasets in \cref{tab:monash_full}.}
    \end{subfigure}%
    ~ 
    \begin{subfigure}[t]{0.5\textwidth}
        \centering
        \includegraphics[width=\columnwidth]{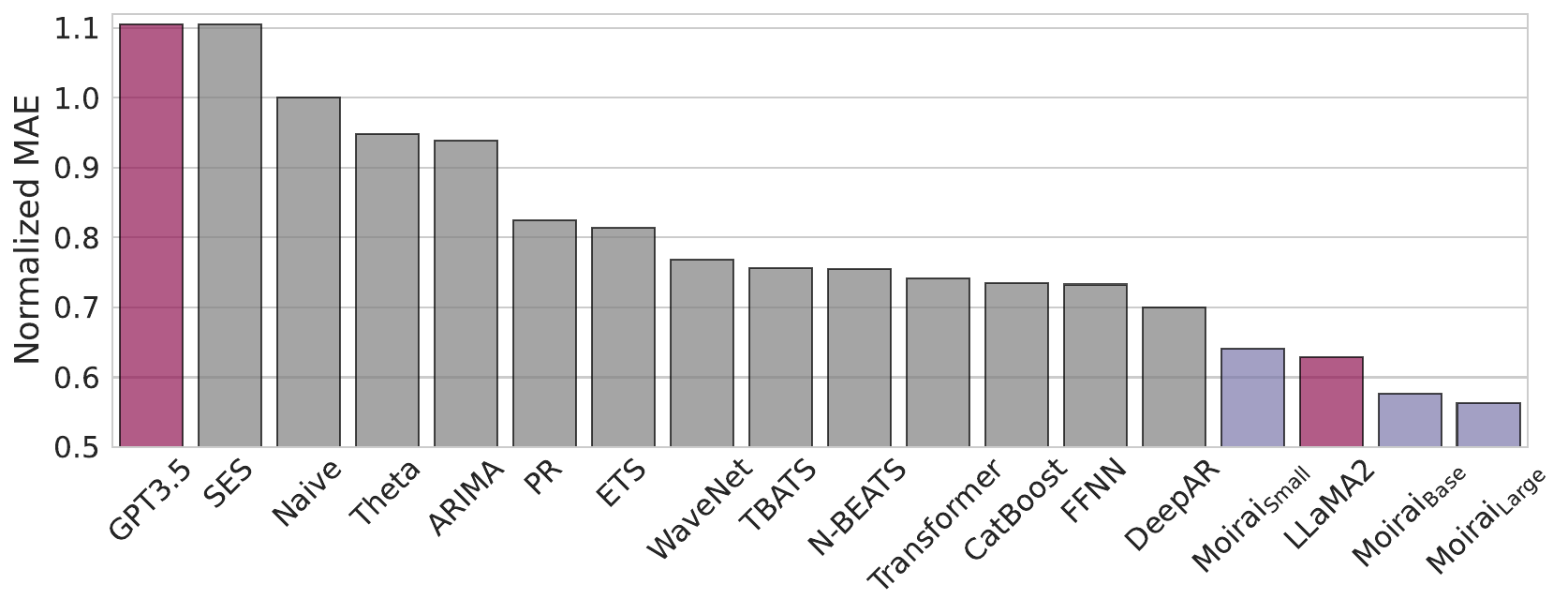}
        \caption{Results aggregated over LLaMA2 subset in \cref{tab:monash_full}.}
    \end{subfigure}
\caption{
Extended aggregate results of the Monash Time Series Forecasting Benchmark as per \cref{fig:monash_summary}. GPT3.5 is our reproduction of LLMTime based on the GPT3.5 API, whereas LLaMA2 is based on the results reported in \citet{gruver2023llmtime}.
}
\label{fig:monash_with_llmtime}
\vskip -0.1in
\end{figure}

%% file: table/monash_full.tex
\begin{table}[htp]
  \centering
  \caption{Full results of Monash Time Series Forecasting Benchmark. MAE is reported.}
  \resizebox{\columnwidth}{!}{
    \begin{tabular}{lcccccccccccccccccc}
    \toprule
          & \textbf{{\smallmodel}} & \textbf{{\basemodel}} & \textbf{{\largemodel}} & \textbf{Naive} & \textbf{SES} & \textbf{Theta} & \textbf{TBATS} & \textbf{ETS} & \textbf{(DHR-)ARIMA} & \textbf{PR} & \textbf{CatBoost} & \textbf{FFNN} & \textbf{DeepAR} & \textbf{N-BEATS} & \textbf{WaveNet} & \textbf{Transformer} & \textbf{GPT3.5} & \textbf{LLaMA2} \\
    \midrule
    M1 Monthly & 2,082.26 & 2,068.63 & 1,983.18 & 2,707.75 & 2,259.04 & 2,166.18 & 2,237.50 & 1,905.28 & 2,080.13 & 2,088.25 & 2,052.32 & 2,162.58 & 1,860.81 & 1,820.37 & 2,184.42 & 2,723.88 & 2562.84 & - \\
    M3 Monthly & 713.41 & 658.17 & 664.03 & 837.14 & 743.41 & 623.71 & 630.59 & 626.46 & 654.8 & 692.97 & 732   & 692.48 & 728.81 & 648.6 & 699.3 & 798.38 & 877.97 & - \\
    M3 Other & 263.54 & 198.62 & 202.41 & 278.43 & 277.83 & 215.35 & 189.42 & 194.98 & 193.02 & 234.43 & 318.13 & 240.17 & 247.56 & 221.85 & 245.29 & 239.24 & 300.30 & - \\
    M4 Monthly & 597.6 & 592.09 & 584.36 & 671.27 & 625.24 & 563.58 & 589.52 & 582.6 & 575.36 & 596.19 & 611.69 & 612.52 & 615.22 & 578.48 & 655.51 & 780.47 & 728.27 & - \\
    M4 Weekly & 339.76 & 328.08 & 301.52 & 347.99 & 336.82 & 333.32 & 296.15 & 335.66 & 321.61 & 293.21 & 364.65 & 338.37 & 351.78 & 277.73 & 359.46 & 378.89 & 518.44 & - \\
    M4 Daily & 189.1 & 192.66 & 189.78 & 180.83 & 178.27 & 178.86 & 176.6 & 193.26 & 179.67 & 181.92 & 231.36 & 177.91 & 299.79 & 190.44 & 189.47 & 201.08 & 266.52 & - \\
    M4 Hourly & 268.04 & 209.87 & 197.79 & 1,218.06 & 1,218.06 & 1,220.97 & 386.27 & 3,358.10 & 1,310.85 & 257.39 & 285.35 & 385.49 & 886.02 & 425.75 & 393.63 & 320.54 & 576.06 & - \\
    Tourism Quarterly & 18,352.44 & 17,196.86 & 15,820.02 & 15,845.10 & 15,014.19 & 7,656.49 & 9,972.42 & 8,925.52 & 10,475.47 & 9,092.58 & 10,267.97 & 8,981.04 & 9,511.37 & 8,640.56 & 9,137.12 & 9,521.67 & 16918.86 & 9311.98 \\
    Tourism Monthly & 3,569.85 & 2,862.06 & 2,688.55 & 5,636.83 & 5,302.10 & 2,069.96 & 2,940.08 & 2,004.51 & 2,536.77 & 2,187.28 & 2,537.04 & 2,022.21 & 1,871.69 & 2,003.02 & 2,095.13 & 2,146.98 & 5608.61 & 3145.48 \\
    CIF 2016 & 655,888.58 & 539,222.03 & 695,156.92 & 578,596.53 & 581,875.97 & 714,818.58 & 855,578.40 & 642,421.42 & 469,059.49 & 563,205.57 & 603,551.30 & 1,495,923.44 & 3,200,418.00 & 679,034.80 & 5,998,224.62 & 4,057,973.00 & 599313.84 & 684057.87 \\
    Aus. Elec. Demand & 266.57 & 201.39 & 177.68 & 659.6 & 659.6 & 665.04 & 370.74 & 1,282.99 & 1,045.92 & 247.18 & 241.77 & 258.76 & 302.41 & 213.83 & 227.5 & 231.45 & 760.81 & 560.48 \\
    Bitcoin & 1.76E+18 & 1.62E+18 & 1.87E+18 & 7.78E+17 & 5.33E+18 & 5.33E+18 & 9.90E+17 & 1.10E+18 & 3.62E+18 & 6.66E+17 & 1.93E+18 & 1.45E+18 & 1.95E+18 & 1.06E+18 & 2.46E+18 & 2.61E+18 & 1.74E+18 & 8.57E+17 \\
    Pedestrian Counts & 54.88 & 54.08 & 41.66 & 170.88 & 170.87 & 170.94 & 222.38 & 216.5 & 635.16 & 44.18 & 43.41 & 46.41 & 44.78 & 66.84 & 46.46 & 47.29 & 97.77 & 65.92 \\
    Vehicle Trips & 24.46 & 23.17 & 21.85 & 31.42 & 29.98 & 30.76 & 21.21 & 30.95 & 30.07 & 27.24 & 22.61 & 22.93 & 22    & 28.16 & 24.15 & 28.01 & 31.48 & - \\
    KDD cup & 39.81 & 38.66 & 39.09 & 42.13 & 42.04 & 42.06 & 39.2  & 44.88 & 52.2  & 36.85 & 34.82 & 37.16 & 48.98 & 49.1  & 37.08 & 44.46 & 42.72 & - \\
    Weather & 1.96  & 1.8   & 1.75  & 2.36  & 2.24  & 2.51  & 2.3   & 2.35  & 2.45  & 8.17  & 2.51  & 2.09  & 2.02  & 2.34  & 2.29  & 2.03  & 2.17  & 2.09 \\
    NN5 Daily & 5.37  & 4.26  & 3.77  & 8.26  & 6.63  & 3.8   & 3.7   & 3.72  & 4.41  & 5.47  & 4.22  & 4.06  & 3.94  & 4.92  & 3.97  & 4.16  & 7.10  & 6.67 \\
    NN5 Weekly & 15.07 & 16.42 & 15.3  & 16.71 & 15.66 & 15.3  & 14.98 & 15.7  & 15.38 & 14.94 & 15.29 & 15.02 & 14.69 & 14.19 & 19.34 & 20.34 & 15.76 & 15.60 \\
    Carparts & 0.53  & 0.47  & 0.49  & 0.65  & 0.55  & 0.53  & 0.58  & 0.56  & 0.56  & 0.41  & 0.53  & 0.39  & 0.39  & 0.98  & 0.4   & 0.39  & 0.44  & - \\
    FRED-MD & 2,568.48 & 2,679.29 & 2,792.55 & 2,825.67 & 2,798.22 & 3,492.84 & 1,989.97 & 2,041.42 & 2,957.11 & 8,921.94 & 2,475.68 & 2,339.57 & 4,264.36 & 2,557.80 & 2,508.40 & 4,666.04 & 2804.64 & 1781.41 \\
    Traffic Hourly & 0.02  & 0.02  & 0.01  & 0.03  & 0.03  & 0.03  & 0.04  & 0.03  & 0.04  & 0.02  & 0.02  & 0.01  & 0.01  & 0.02  & 0.02  & 0.01  & 0.03  & 0.02 \\
    Traffic Weekly & 1.17  & 1.14  & 1.13  & 1.19  & 1.12  & 1.13  & 1.17  & 1.14  & 1.22  & 1.13  & 1.17  & 1.15  & 1.18  & 1.11  & 1.2   & 1.42  & 1.15  & 1.15 \\
    Rideshare & 1.35  & 1.39  & 1.29  & 6.29  & 6.29  & 7.62  & 6.45  & 6.29  & 3.37  & 6.3   & 6.07  & 6.59  & 6.28  & 5.55  & 2.75  & 6.29  & 6.28  & - \\
    Hospital & 23    & 19.4  & 19.44 & 24.07 & 21.76 & 18.54 & 17.43 & 17.97 & 19.6  & 19.24 & 19.17 & 22.86 & 18.25 & 20.18 & 19.35 & 36.19 & 25.68 & 22.75 \\
    COVID Deaths & 124.32 & 126.11 & 117.11 & 353.71 & 353.71 & 321.32 & 96.29 & 85.59 & 85.77 & 347.98 & 475.15 & 144.14 & 201.98 & 158.81 & 1,049.48 & 408.66 & 653.31 & 66.14 \\
    Temperature Rain & 5.3   & 5.08  & 5.27  & 9.39  & 8.18  & 8.22  & 7.14  & 8.21  & 7.19  & 6.13  & 6.76  & 5.56  & 5.37  & 7.28  & 5.81  & 5.24  & 6.37  & - \\
    Sunspot & 0.11  & 0.08  & 0.13  & 3.93  & 4.93  & 4.93  & 2.57  & 4.93  & 2.57  & 3.83  & 2.27  & 7.97  & 0.77  & 14.47 & 0.17  & 0.13  & 5.07  & 0.28 \\
    Saugeen River Flow & 24.07 & 24.4  & 24.76 & 21.5  & 21.5  & 21.49 & 22.26 & 30.69 & 22.38 & 25.24 & 21.28 & 22.98 & 23.51 & 27.92 & 22.17 & 28.06 & 34.84 & 23.01 \\
    US Births & 872.51 & 624.3 & 476.5 & 1,152.67 & 1,192.20 & 586.93 & 399   & 419.73 & 526.33 & 574.93 & 441.7 & 557.87 & 424.93 & 422   & 504.4 & 452.87 & 1374.99 & 638.82 \\
    \bottomrule
    \end{tabular}%
    }
  \label{tab:monash_full}%
\end{table}%

%% file: table/pf_full.tex
\begin{table}[htp]
  \centering
  \caption{Full results for probabilistic forecasting experiments. Best results are highlighted in \textbf{bold}, and second best results are \underline{underlined}.}
  \resizebox{\columnwidth}{!}{
    \begin{tabular}{ccccccccccc}
    \toprule
    &       & \multicolumn{3}{c}{\textbf{Zero-shot}} & \multicolumn{4}{c}{\textbf{Full-shot}} & \multicolumn{2}{c}{\textbf{Baseline}} \\
    \cmidrule(lr){3-5} \cmidrule(lr){6-9} \cmidrule(lr){10-11}
          &       & \textbf{{\smallmodel}} & \textbf{{\basemodel}} & \textbf{{\largemodel}} & \textbf{PatchTST} & \textbf{TiDE} & \textbf{TFT} & \textbf{DeepAR} & \textbf{AutoARIMA} & \textbf{Seasonal Naive} \\
    \midrule
    \multirow{6}[2]{*}{Electricity} & \textbf{CRPS} & 0.072 & 0.055 & \underline{0.050} & 0.052{\scriptsize$\pm$}0.00 & \boldmath{}\textbf{0.048{\scriptsize$\pm$}0.00}\unboldmath{} & 0.050{\scriptsize$\pm$}0.00 & 0.065{\scriptsize$\pm$}0.01 & 0.327 & 0.070 \\
          & \textbf{MSIS} & 7.999 & 6.172 & 5.875 & \underline{5.744{\scriptsize$\pm$}0.12} & \boldmath{}\textbf{5.672{\scriptsize$\pm$}0.08}\unboldmath{} & 6.278{\scriptsize$\pm$}0.24 & 6.893{\scriptsize$\pm$}0.82 & 29.412 & 35.251 \\
          & \textbf{sMAPE} & 0.134 & 0.111 & 0.106 & 0.107{\scriptsize$\pm$}0.00 & \boldmath{}\textbf{0.102{\scriptsize$\pm$}0.00}\unboldmath{} & \underline{0.106{\scriptsize$\pm$}0.01} & 0.118{\scriptsize$\pm$}0.02 & 0.318 & 0.108 \\
          & \textbf{MASE} & 0.981 & 0.792 & 0.751 & 0.753{\scriptsize$\pm$}0.01 & \boldmath{}\textbf{0.706{\scriptsize$\pm$}0.02}\unboldmath{} & \underline{0.747{\scriptsize$\pm$}0.03} & 0.844{\scriptsize$\pm$}0.16 & 3.229 & 0.881 \\
          & \textbf{ND} & 0.092 & 0.069 & 0.063 & 0.065{\scriptsize$\pm$}0.00 & \boldmath{}\textbf{0.061{\scriptsize$\pm$}0.00}\unboldmath{} & \underline{0.063{\scriptsize$\pm$}0.00} & 0.080{\scriptsize$\pm$}0.02 & 0.357 & 0.070 \\
          & \textbf{NRMSE} & 0.840 & 0.551 & \textbf{0.465} & 0.506{\scriptsize$\pm$}0.02 & 0.514{\scriptsize$\pm$}0.02 & 0.511{\scriptsize$\pm$}0.02 & 0.704{\scriptsize$\pm$}0.11 & 3.296 & \underline{0.478} \\
    \midrule
    \multirow{6}[2]{*}{Solar} & \textbf{CRPS} & 0.471 & \underline{0.419} & \textbf{0.406} & 0.518{\scriptsize$\pm$}0.09 & 0.420{\scriptsize$\pm$}0.00 & 0.446{\scriptsize$\pm$}0.03 & 0.431{\scriptsize$\pm$}0.01 & 1.055 & 0.512 \\
          & \textbf{MSIS} & 8.425 & \underline{7.011} & \textbf{6.250} & 8.447{\scriptsize$\pm$}1.59 & 13.754{\scriptsize$\pm$}0.32 & 8.057{\scriptsize$\pm$}3.51 & 11.181{\scriptsize$\pm$}0.67 & 25.849 & 48.130 \\
          & \textbf{sMAPE} & 1.445 & 1.410 & 1.400 & 1.501{\scriptsize$\pm$}0.10 & 1.400{\scriptsize$\pm$}0.00 & 1.391{\scriptsize$\pm$}0.01 & \underline{1.385{\scriptsize$\pm$}0.00} & 1.685 & \textbf{0.691} \\
          & \textbf{MASE} & 1.465 & 1.292 & 1.237 & 1.607{\scriptsize$\pm$}0.25 & 1.265{\scriptsize$\pm$}0.02 & 1.399{\scriptsize$\pm$}0.11 & \underline{1.222{\scriptsize$\pm$}0.01} & 2.583 & \textbf{1.203} \\
          & \textbf{ND} & 0.624 & 0.551 & 0.528 & 0.685{\scriptsize$\pm$}0.11 & 0.538{\scriptsize$\pm$}0.01 & 0.594{\scriptsize$\pm$}0.05 & \underline{0.520{\scriptsize$\pm$}0.00} & 1.098 & \textbf{0.512} \\
          & \textbf{NRMSE} & 1.135 & 1.034 & \textbf{1.014} & 1.408{\scriptsize$\pm$}0.26 & 1.093{\scriptsize$\pm$}0.00 & 1.236{\scriptsize$\pm$}0.06 & \underline{1.033{\scriptsize$\pm$}0.01} & 1.784 & 1.168 \\
    \midrule
    \multirow{6}[2]{*}{Walmart} & \textbf{CRPS} & 0.103 & 0.093 & 0.098 & \underline{0.082{\scriptsize$\pm$}0.01} & \boldmath{}\textbf{0.077{\scriptsize$\pm$}0.00}\unboldmath{} & 0.087{\scriptsize$\pm$}0.00 & 0.121{\scriptsize$\pm$}0.00 & 0.124 & 0.151 \\
          & \textbf{MSIS} & 9.371 & 8.421 & 8.520 & \boldmath{}\textbf{6.005{\scriptsize$\pm$}0.21}\unboldmath{} & \underline{6.258{\scriptsize$\pm$}0.12} & 8.718{\scriptsize$\pm$}0.10 & 12.502{\scriptsize$\pm$}0.03 & 9.888 & 49.458 \\
          & \textbf{sMAPE} & 0.179 & 0.168 & 0.174 & \underline{0.150{\scriptsize$\pm$}0.01} & \boldmath{}\textbf{0.145{\scriptsize$\pm$}0.00}\unboldmath{} & 0.172{\scriptsize$\pm$}0.00 & 0.216{\scriptsize$\pm$}0.00 & 0.219 & 0.205 \\
          & \textbf{MASE} & 1.048 & 0.964 & 1.007 & \underline{0.867{\scriptsize$\pm$}0.09} & \boldmath{}\textbf{0.814{\scriptsize$\pm$}0.01}\unboldmath{} & 0.948{\scriptsize$\pm$}0.02 & 1.193{\scriptsize$\pm$}0.02 & 1.131 & 1.236 \\
          & \textbf{ND} & 0.129 & 0.117 & 0.124 & \underline{0.105{\scriptsize$\pm$}0.01} & \boldmath{}\textbf{0.097{\scriptsize$\pm$}0.00}\unboldmath{} & 0.108{\scriptsize$\pm$}0.00 & 0.147{\scriptsize$\pm$}0.00 & 0.141 & 0.151 \\
          & \textbf{NRMSE} & 0.324 & 0.291 & 0.332 & \underline{0.218{\scriptsize$\pm$}0.02} & \boldmath{}\textbf{0.204{\scriptsize$\pm$}0.00}\unboldmath{} & 0.235{\scriptsize$\pm$}0.01 & 0.298{\scriptsize$\pm$}0.00 & 0.305 & 0.328 \\
    \midrule
    \multirow{6}[2]{*}{Weather} & \textbf{CRPS} & 0.049 & \textbf{0.041} & 0.051 & 0.059{\scriptsize$\pm$}0.01 & 0.054{\scriptsize$\pm$}0.00 & \underline{0.043{\scriptsize$\pm$}0.00} & 0.132{\scriptsize$\pm$}0.11 & 0.252 & 0.068 \\
          & \textbf{MSIS} & 5.236 & \underline{5.136} & \textbf{4.962} & 7.759{\scriptsize$\pm$}0.49 & 8.095{\scriptsize$\pm$}1.74 & 7.791{\scriptsize$\pm$}0.44 & 21.651{\scriptsize$\pm$}17.34 & 19.805 & 31.293 \\
          & \textbf{sMAPE} & 0.686 & \underline{0.623} & 0.688 & 0.668{\scriptsize$\pm$}0.01 & 0.636{\scriptsize$\pm$}0.01 & 0.672{\scriptsize$\pm$}0.01 & 0.776{\scriptsize$\pm$}0.05 & 0.770 & \textbf{0.401} \\
          & \textbf{MASE} & 0.521 & \textbf{0.487} & \underline{0.515} & 0.844{\scriptsize$\pm$}0.19 & 0.832{\scriptsize$\pm$}0.13 & 0.692{\scriptsize$\pm$}0.02 & 3.170{\scriptsize$\pm$}3.47 & 0.938 & 0.782 \\
          & \textbf{ND} & 0.063 & \textbf{0.048} & 0.063 & 0.072{\scriptsize$\pm$}0.01 & 0.066{\scriptsize$\pm$}0.01 & \underline{0.051{\scriptsize$\pm$}0.00} & 0.163{\scriptsize$\pm$}0.15 & 0.139 & 0.068 \\
          & \textbf{NRMSE} & 0.229 & 0.417 & 0.331 & 0.260{\scriptsize$\pm$}0.01 & \underline{0.214{\scriptsize$\pm$}0.00} & \boldmath{}\textbf{0.211{\scriptsize$\pm$}0.00}\unboldmath{} & 0.486{\scriptsize$\pm$}0.43 & 0.465 & 0.290 \\
    \midrule
    \multirow{6}[2]{*}{Istanbul Traffic} & \textbf{CRPS} & 0.173 & 0.116 & 0.112 & 0.112{\scriptsize$\pm$}0.00 & 0.110{\scriptsize$\pm$}0.01 & \underline{0.110{\scriptsize$\pm$}0.01} & \boldmath{}\textbf{0.108{\scriptsize$\pm$}0.00}\unboldmath{} & 0.589 & 0.257 \\
          & \textbf{MSIS} & 5.937 & 4.461 & 4.277 & \boldmath{}\textbf{3.813{\scriptsize$\pm$}0.09}\unboldmath{} & 4.752{\scriptsize$\pm$}0.17 & \underline{4.057{\scriptsize$\pm$}0.44} & 4.094{\scriptsize$\pm$}0.31 & 16.317 & 45.473 \\
          & \textbf{sMAPE} & 0.359 & 0.284 & 0.288 & 0.287{\scriptsize$\pm$}0.01 & \underline{0.280{\scriptsize$\pm$}0.01} & 0.287{\scriptsize$\pm$}0.01 & \boldmath{}\textbf{0.249{\scriptsize$\pm$}0.01}\unboldmath{} & 1.141 & 0.391 \\
          & \textbf{MASE} & 0.990 & 0.644 & 0.631 & 0.653{\scriptsize$\pm$}0.02 & \underline{0.618{\scriptsize$\pm$}0.03} & 0.620{\scriptsize$\pm$}0.03 & \boldmath{}\textbf{0.613{\scriptsize$\pm$}0.03}\unboldmath{} & 3.358 & 1.137 \\
          & \textbf{ND} & 0.224 & 0.146 & 0.143 & 0.148{\scriptsize$\pm$}0.01 & \underline{0.140{\scriptsize$\pm$}0.01} & 0.141{\scriptsize$\pm$}0.01 & \boldmath{}\textbf{0.139{\scriptsize$\pm$}0.01}\unboldmath{} & 0.758 & 0.257 \\
          & \textbf{NRMSE} & 0.294 & 0.194 & 0.186 & 0.190{\scriptsize$\pm$}0.01 & 0.185{\scriptsize$\pm$}0.01 & \underline{0.185{\scriptsize$\pm$}0.01} & \boldmath{}\textbf{0.181{\scriptsize$\pm$}0.01}\unboldmath{} & 0.959 & 0.384 \\
    \midrule
    \multirow{6}[2]{*}{Turkey Power} & \textbf{CRPS} & 0.048 & 0.040 & \textbf{0.036} & 0.054{\scriptsize$\pm$}0.01 & 0.046{\scriptsize$\pm$}0.01 & \underline{0.039{\scriptsize$\pm$}0.00} & 0.066{\scriptsize$\pm$}0.02 & 0.116 & 0.085 \\
          & \textbf{MSIS} & 7.127 & \underline{6.766} & \textbf{6.341} & 8.978{\scriptsize$\pm$}0.51 & 8.579{\scriptsize$\pm$}0.52 & 7.943{\scriptsize$\pm$}0.31 & 13.520{\scriptsize$\pm$}1.17 & 14.863 & 36.256 \\
          & \textbf{sMAPE} & 0.389 & 0.378 & 0.375 & 0.416{\scriptsize$\pm$}0.01 & 0.389{\scriptsize$\pm$}0.00 & 0.383{\scriptsize$\pm$}0.00 & 0.404{\scriptsize$\pm$}0.01 & \underline{0.244} & \textbf{0.125} \\
          & \textbf{MASE} & 0.948 & \underline{0.888} & \textbf{0.870} & 1.234{\scriptsize$\pm$}0.12 & 0.904{\scriptsize$\pm$}0.02 & 0.890{\scriptsize$\pm$}0.05 & 1.395{\scriptsize$\pm$}0.30 & 1.700 & 0.906 \\
          & \textbf{ND} & 0.061 & 0.051 & \textbf{0.046} & 0.071{\scriptsize$\pm$}0.01 & 0.059{\scriptsize$\pm$}0.01 & \underline{0.049{\scriptsize$\pm$}0.00} & 0.083{\scriptsize$\pm$}0.02 & 0.150 & 0.085 \\
          & \textbf{NRMSE} & 0.149 & 0.118 & \textbf{0.102} & 0.158{\scriptsize$\pm$}0.01 & 0.139{\scriptsize$\pm$}0.03 & \underline{0.104{\scriptsize$\pm$}0.01} & 0.181{\scriptsize$\pm$}0.05 & 0.383 & 0.231 \\
    \bottomrule
    \end{tabular}%
  }
  \label{tab:pf_full}%
\end{table}%

%% file: table/lsf_full.tex
\begin{table}[htp]
  \centering
  \caption{Full results of long sequence forecasting experiments. Best results are highlighted in \textbf{bold}, and second best results are \underline{underlined}. Full-shot results are obtained from \citet{liu2023itransformer}.}
  \resizebox{\columnwidth}{!}{
    \begin{tabular}{crcccccccccccccccccccccc}
          \toprule
          &       & \multicolumn{6}{c}{\textbf{Zero-shot}}                 & \multicolumn{16}{c}{\textbf{Full-shot}} \\
          \cmidrule(lr){3-8} \cmidrule(lr){9-24}
          &       & \multicolumn{2}{c}{\textbf{{\smallmodel}}} & \multicolumn{2}{c}{\textbf{{\basemodel}}} & \multicolumn{2}{c}{\textbf{{\largemodel}}} & \multicolumn{2}{c}{\textbf{iTransformer}} & \multicolumn{2}{c}{\textbf{TimesNet}} & \multicolumn{2}{c}{\textbf{PatchTST}} & \multicolumn{2}{c}{\textbf{Crossformer}} & \multicolumn{2}{c}{\textbf{TiDE}} & \multicolumn{2}{c}{\textbf{DLinear}} & \multicolumn{2}{c}{\textbf{SCINet}} & \multicolumn{2}{c}{\textbf{FEDformer}} \\
          \midrule
          &       & \textbf{MSE} & \textbf{MAE} & \textbf{MSE} & \textbf{MAE} & \textbf{MSE} & \textbf{MAE} & \textbf{MSE} & \textbf{MAE} & \textbf{MSE} & \textbf{MAE} & \textbf{MSE} & \textbf{MAE} & \textbf{MSE} & \textbf{MAE} & \textbf{MSE} & \textbf{MAE} & \textbf{MSE} & \textbf{MAE} & \textbf{MSE} & \textbf{MAE} & \textbf{MSE} & \textbf{MAE} \\
          \midrule    \multirow{4}[1]{*}{ETTh1} & 96    & \textbf{0.375} & 0.402 & 0.384 & 0.402 & 0.380 & 0.398 & 0.386 & 0.405 & 0.384 & 0.402 & 0.414 & 0.419 & 0.423 & 0.448 & 0.479 & 0.464 & 0.386 & 0.400 & 0.654 & 0.599 & \underline{0.376} & 0.419 \\
          & 192   & \textbf{0.399} & \textbf{0.419} & 0.425 & 0.429 & 0.440 & 0.434 & 0.441 & 0.436 & 0.436 & 0.429 & 0.460 & 0.445 & 0.471 & 0.474 & 0.525 & 0.492 & 0.437 & 0.432 & 0.719 & 0.631 & 0.420 & 0.448 \\
          & 336   & \textbf{0.412} & \textbf{0.429} & 0.456 & \underline{0.450} & 0.514 & 0.474 & 0.487 & 0.458 & 0.491 & 0.469 & 0.501 & 0.466 & 0.570 & 0.546 & 0.565 & 0.515 & 0.481 & 0.459 & 0.778 & 0.659 & 0.459 & 0.465 \\
          & 720   & \textbf{0.413} & \textbf{0.444} & 0.470 & 0.473 & 0.705 & 0.568 & 0.503 & 0.491 & 0.521 & 0.500 & 0.500 & 0.488 & 0.653 & 0.621 & 0.594 & 0.558 & 0.519 & 0.516 & 0.836 & 0.699 & 0.506 & 0.507 \\
    \midrule
    \multirow{4}[0]{*}{ETTh2} & 96    & \underline{0.281} & 0.334 & \textbf{0.277} & 0.327 & 0.287 & 0.325 & 0.297 & 0.349 & 0.340 & 0.374 & 0.302 & 0.348 & 0.745 & 0.584 & 0.400 & 0.440 & 0.333 & 0.387 & 0.707 & 0.621 & 0.358 & 0.397 \\
          & 192   & \underline{0.340} & 0.373 & \textbf{0.340} & 0.374 & 0.347 & 0.367 & 0.380 & 0.400 & 0.402 & 0.414 & 0.388 & 0.400 & 0.877 & 0.656 & 0.528 & 0.509 & 0.477 & 0.476 & 0.860 & 0.689 & 0.429 & 0.439 \\
          & 336   & \textbf{0.362} & 0.393 & \underline{0.371} & 0.401 & 0.377 & 0.393 & 0.428 & 0.432 & 0.452 & 0.541 & 0.426 & 0.433 & 1.043 & 0.731 & 0.643 & 0.571 & 0.594 & 0.541 & 1.000 & 0.744 & 0.496 & 0.487 \\
          & 720   & \textbf{0.380} & 0.416 & \underline{0.394} & 0.426 & 0.404 & 0.421 & 0.427 & 0.445 & 0.462 & 0.657 & 0.431 & 0.446 & 1.104 & 0.763 & 0.874 & 0.679 & 0.831 & 0.657 & 1.249 & 0.838 & 0.463 & 0.474 \\
    \midrule
    \multirow{4}[0]{*}{ETTm1} & 96    & 0.404 & 0.383 & 0.335 & 0.360 & 0.353 & 0.363 & \underline{0.334} & 0.368 & 0.338 & 0.375 & \textbf{0.329} & 0.367 & 0.404 & 0.426 & 0.364 & 0.387 & 0.345 & 0.372 & 0.418 & 0.438 & 0.379 & 0.419 \\
          & 192   & 0.435 & 0.402 & \textbf{0.366} & 0.379 & 0.376 & 0.380 & 0.377 & 0.391 & 0.374 & 0.387 & \underline{0.367} & 0.385 & 0.450 & 0.451 & 0.398 & 0.404 & 0.380 & 0.389 & 0.439 & 0.450 & 0.426 & 0.441 \\
          & 336   & 0.462 & 0.416 & \textbf{0.391} & \underline{0.394} & 0.399 & 0.395 & 0.426 & 0.420 & 0.410 & 0.411 & 0.399 & 0.410 & 0.532 & 0.515 & 0.428 & 0.425 & 0.413 & 0.413 & 0.490 & 0.485 & 0.445 & 0.459 \\
          & 720   & 0.490 & 0.437 & 0.434 & \underline{0.419} & 0.432 & \textbf{0.417} & 0.491 & 0.459 & 0.478 & 0.450 & 0.454 & 0.439 & 0.666 & 0.589 & 0.487 & 0.461 & 0.474 & 0.453 & 0.595 & 0.550 & 0.543 & 0.490 \\
    \midrule
    \multirow{4}[0]{*}{ETTm2} & 96    & 0.205 & 0.282 & 0.195 & 0.269 & 0.189 & 0.260 & \underline{0.180} & 0.264 & 0.187 & 0.267 & \textbf{0.175} & 0.259 & 0.287 & 0.366 & 0.207 & 0.305 & 0.193 & 0.292 & 0.286 & 0.377 & 0.203 & 0.287 \\
          & 192   & 0.261 & 0.318 & \underline{0.247} & 0.303 & 0.247 & 0.300 & 0.250 & 0.309 & 0.249 & 0.309 & \textbf{0.241} & 0.302 & 0.414 & 0.492 & 0.290 & 0.364 & 0.284 & 0.362 & 0.399 & 0.445 & 0.269 & 0.328 \\
          & 336   & 0.319 & 0.355 & \textbf{0.291} & 0.333 & \underline{0.295} & 0.334 & 0.311 & 0.348 & 0.321 & 0.351 & 0.305 & 0.343 & 0.597 & 0.542 & 0.377 & 0.422 & 0.369 & 0.427 & 0.637 & 0.591 & 0.325 & 0.366 \\
          & 720   & 0.415 & 0.410 & \textbf{0.355} & 0.377 & \underline{0.372} & 0.386 & 0.412 & 0.407 & 0.408 & 0.403 & 0.402 & 0.400 & 1.730 & 1.042 & 0.558 & 0.524 & 0.554 & 0.522 & 0.960 & 0.735 & 0.421 & 0.415 \\
    \midrule
    \multirow{4}[0]{*}{Electricity} & 96    & 0.205 & 0.299 & 0.158 & 0.248 & \underline{0.152} & 0.242 & \textbf{0.148} & 0.240 & 0.168 & 0.272 & 0.195 & 0.285 & 0.219 & 0.314 & 0.237 & 0.329 & 0.197 & 0.282 & 0.247 & 0.345 & 0.193 & 0.308 \\
          & 192   & 0.220 & 0.310 & 0.174 & 0.263 & \underline{0.171} & 0.259 & \textbf{0.162} & 0.253 & 0.184 & 0.289 & 0.199 & 0.289 & 0.231 & 0.322 & 0.236 & 0.330 & 0.196 & 0.285 & 0.257 & 0.355 & 0.201 & 0.315 \\
          & 336   & 0.236 & 0.323 & \underline{0.191} & 0.278 & 0.192 & 0.278 & \textbf{0.178} & 0.269 & 0.198 & 0.300 & 0.215 & 0.305 & 0.246 & 0.337 & 0.249 & 0.344 & 0.209 & 0.301 & 0.269 & 0.369 & 0.214 & 0.329 \\
          & 720   & 0.270 & 0.347 & 0.229 & 0.307 & 0.236 & 0.313 & \underline{0.225} & 0.317 & \textbf{0.220} & 0.320 & 0.256 & 0.337 & 0.280 & 0.363 & 0.284 & 0.373 & 0.245 & 0.333 & 0.299 & 0.390 & 0.246 & 0.355 \\
    \midrule
    \multirow{4}[0]{*}{Weather} & 96    & 0.173 & 0.212 & \underline{0.167} & 0.203 & 0.177 & 0.208 & 0.174 & 0.214 & 0.172 & 0.220 & 0.177 & 0.218 & \textbf{0.158} & 0.230 & 0.202 & 0.261 & 0.196 & 0.255 & 0.221 & 0.306 & 0.217 & 0.296 \\
          & 192   & 0.216 & 0.250 & \underline{0.209} & 0.241 & 0.219 & 0.249 & 0.221 & 0.254 & 0.219 & 0.261 & 0.225 & 0.259 & \textbf{0.206} & 0.277 & 0.242 & 0.298 & 0.237 & 0.296 & 0.261 & 0.340 & 0.276 & 0.336 \\
          & 336   & \underline{0.260} & 0.282 & \textbf{0.256} & 0.276 & 0.277 & 0.292 & 0.278 & 0.296 & 0.280 & 0.306 & 0.278 & 0.297 & 0.272 & 0.335 & 0.287 & 0.335 & 0.283 & 0.335 & 0.309 & 0.378 & 0.339 & 0.380 \\
          & 720   & \textbf{0.320} & \underline{0.322} & \underline{0.321} & 0.323 & 0.365 & 0.350 & 0.358 & 0.349 & 0.365 & 0.359 & 0.354 & 0.348 & 0.398 & 0.418 & 0.351 & 0.386 & 0.345 & 0.381 & 0.377 & 0.427 & 0.403 & 0.428 \\
    \bottomrule
    \end{tabular}%
    }
  \label{tab:lsf_full}%
\end{table}%

%% file: table/comp_cost.tex
\begin{table}[ht]
  \centering
  \caption{Computational cost in terms of seconds of various models in terms of seconds for inference for a batch size of 32. ``(32)'' for {\modelname} refers to patch size.}
    \begin{tabular}{ccccccccccc}
    \toprule
          & \multicolumn{5}{c}{Context Length}    & \multicolumn{5}{c}{Prediction Length} \\
          \cmidrule(lr){2-6} \cmidrule(lr){7-11}
          & 1000  & 2000  & 3000  & 4000  & 5000  & 1000  & 2000  & 3000  & 4000  & 5000 \\
    \midrule
    {\smallmodel} (32) & 0.03  & 0.04  & 0.05  & 0.06  & 0.07  & 0.03  & 0.04  & 0.05  & 0.06  & 0.07 \\
    {\basemodel} (32) & 0.05  & 0.06  & 0.08  & 0.11  & 0.13  & 0.05  & 0.06  & 0.08  & 0.11  & 0.13 \\
    {\largemodel} (32) & 0.09  & 0.14  & 0.19  & 0.25  & 0.3   & 0.09  & 0.14  & 0.19  & 0.25  & 0.3 \\
    PatchTST & 0.01  & 0.02  & 0.02  & 0.03  & 0.04  & 0.01  & 0.01  & 0.01  & 0.01  & 0.02 \\
    TiDE  & 0.01  & 0.01  & 0.01  & 0.01  & 0.01  & 0.01  & 0.01  & 0.01  & 0.01  & 0.01 \\
    TFT   & 0.02  & 0.04  & 0.06  & 0.08  & 0.09  & 0.03  & 0.07  & 0.12  & 0.17  & OOM \\
    DeepAR & 0.26  & 0.32  & 0.37  & 0.43  & 0.49  & 2.02  & 4.06  & 6.1   & 8.17  & 10.24 \\
    \bottomrule
    \end{tabular}%
  \label{tab:comp_cost}%
\end{table}%

%% file: appendix/forecast_viz.tex
\newpage
\section{Forecast Visualizations}
\input{figure/forecast_viz}

%% file: figure/forecast_viz.tex
\begin{figure}[htp]
\begin{center}
    \begin{subfigure}[t]{\columnwidth}
        \centering
        \includegraphics[width=\columnwidth]{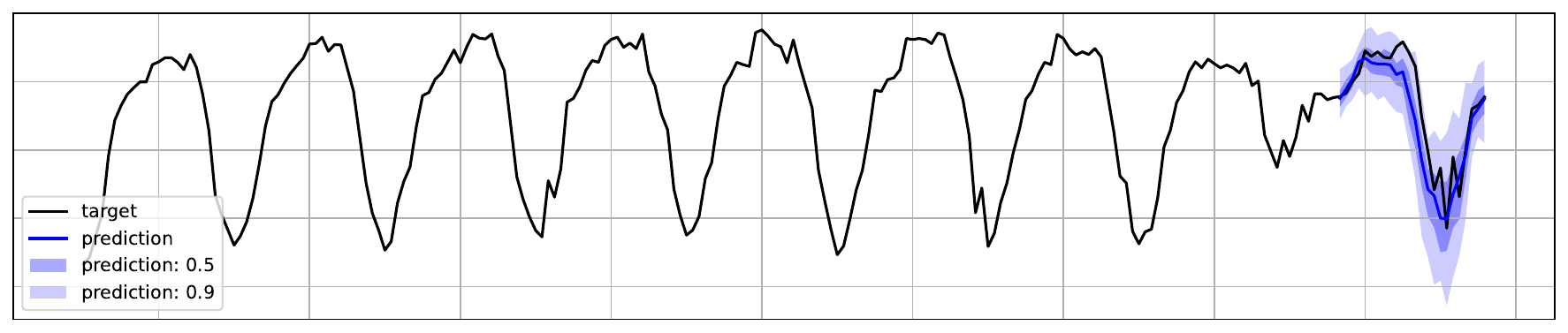}
        \caption{ETTh1-1}
    \end{subfigure}
        \begin{subfigure}[t]{\columnwidth}
        \centering
        \includegraphics[width=\columnwidth]{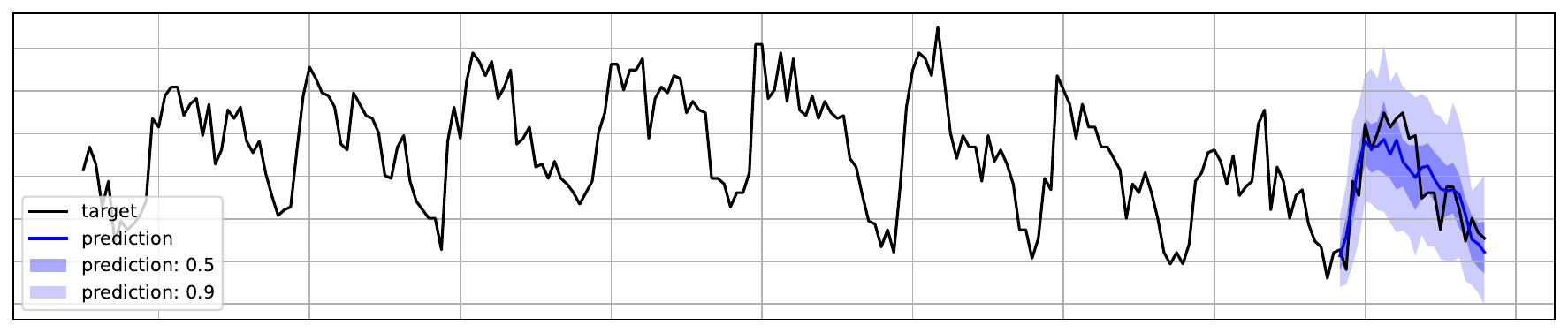}
        \caption{ETTh1-2}
    \end{subfigure}
    \begin{subfigure}[t]{\columnwidth}
        \centering
        \includegraphics[width=\columnwidth]{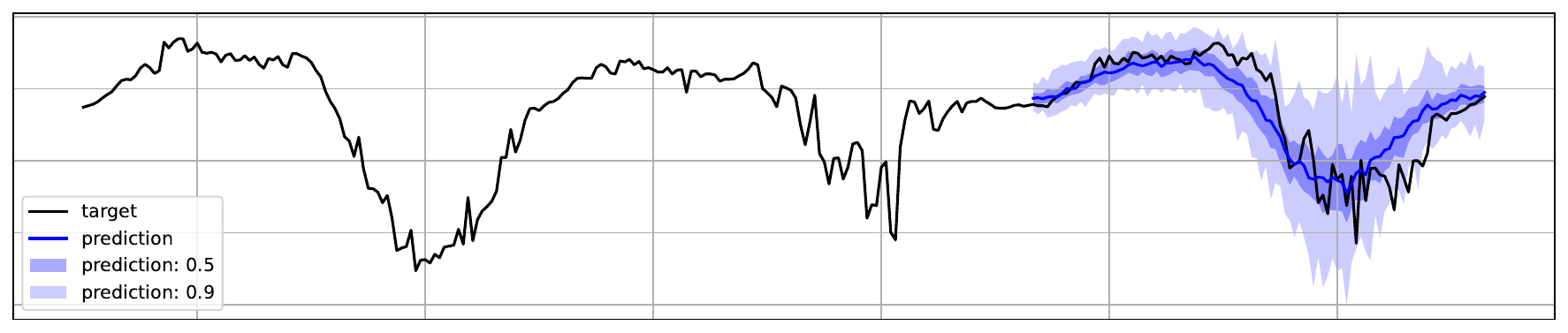}
        \caption{ETTm1-1}
    \end{subfigure}
        \begin{subfigure}[t]{\columnwidth}
        \centering
        \includegraphics[width=\columnwidth]{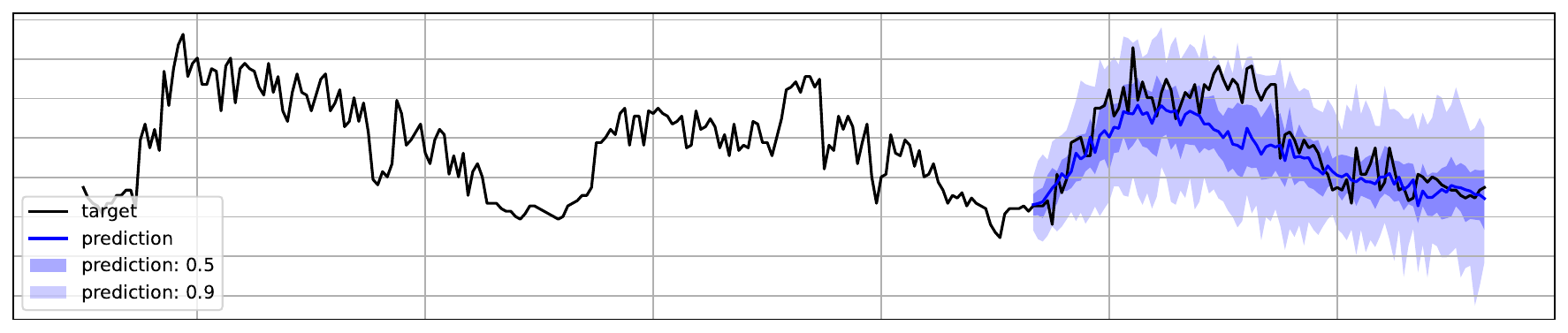}
        \caption{ETTm1-2}
    \end{subfigure}
\caption{Visualizations of \textbf{zero-shot forecasts} from {\basemodel} on ETTh1 and ETTm1 datasets.}
\label{fig:viz1}
\end{center}
\end{figure}

\begin{figure}[htp]
\begin{center}
    \begin{subfigure}[t]{\columnwidth}
        \centering
        \includegraphics[width=\columnwidth]{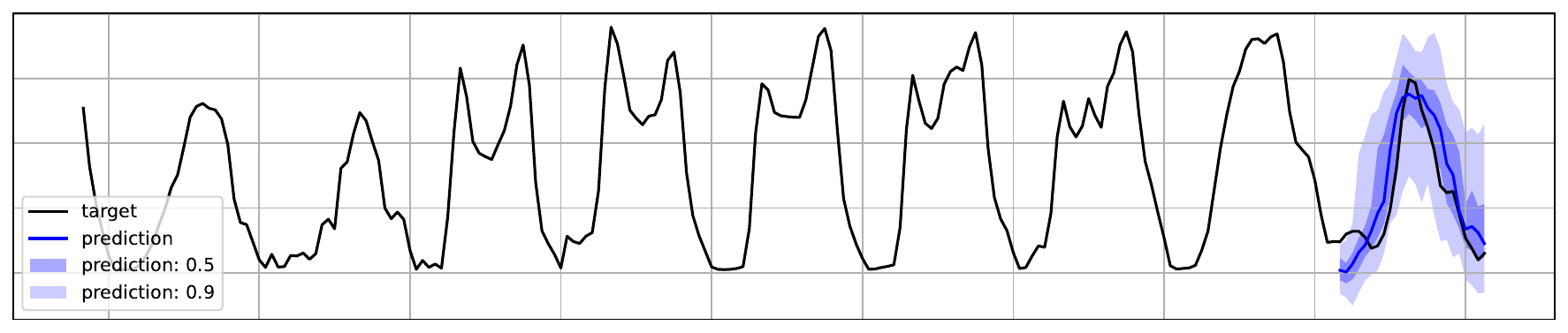}
        \caption{Istanbul Traffic-1}
    \end{subfigure}
        \begin{subfigure}[t]{\columnwidth}
        \centering
        \includegraphics[width=\columnwidth]{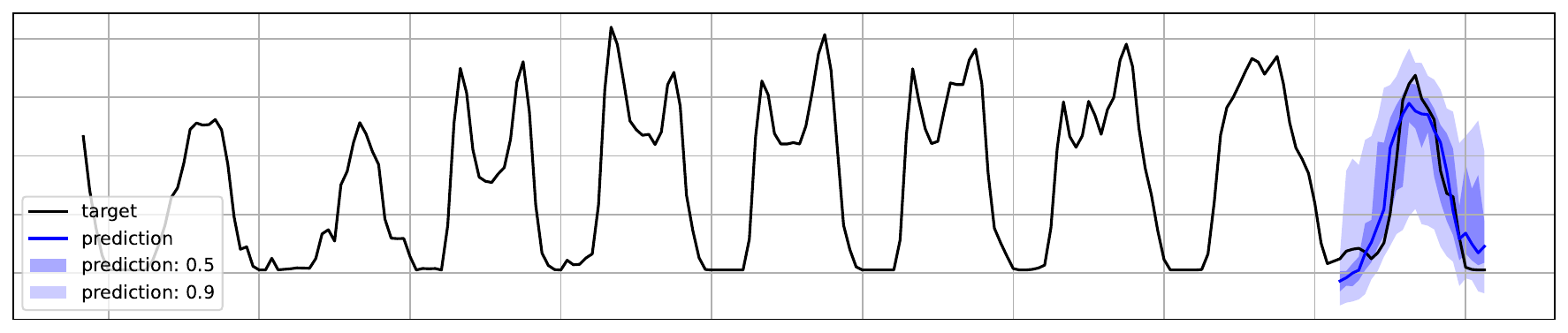}
        \caption{Istanbul Traffic-2}
    \end{subfigure}
    \begin{subfigure}[t]{\columnwidth}
        \centering
        \includegraphics[width=\columnwidth]{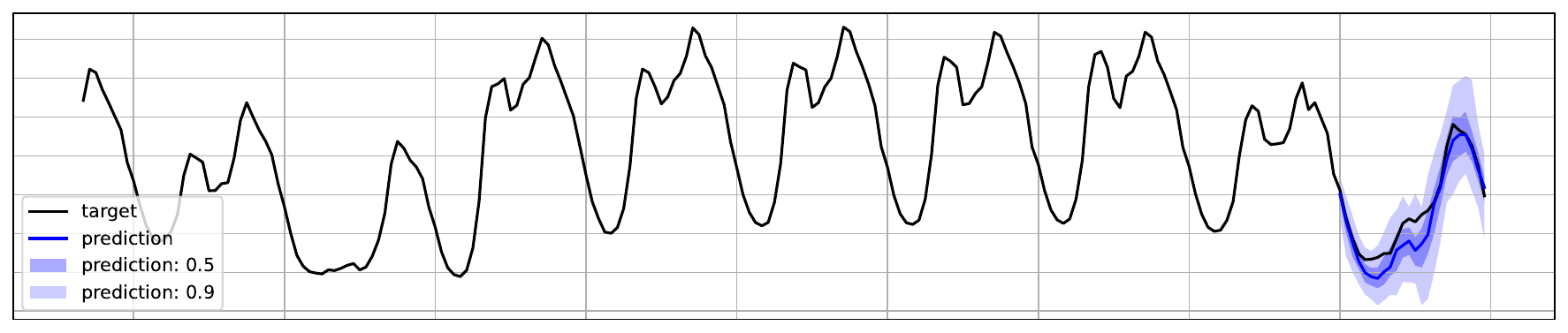}
        \caption{Turkey Power-1}
    \end{subfigure}
        \begin{subfigure}[t]{\columnwidth}
        \centering
        \includegraphics[width=\columnwidth]{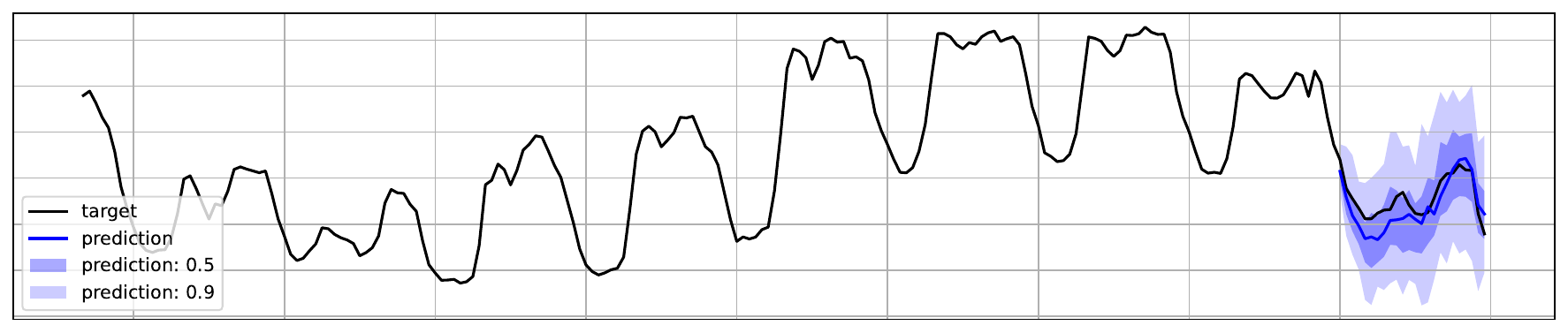}
        \caption{Turkey Power-2}
    \end{subfigure}
\caption{Visualizations of \textbf{zero-shot forecasts} from {\basemodel} on Istanbul Traffic and Turkey Power datasets.}
\label{fig:viz2}
\end{center}
\end{figure}